\documentclass[10pt, a4paper]{article}

\usepackage[final]{lrec2026}

\usepackage[whole]{bxcjkjatype}
\usepackage{tcolorbox}
\usepackage{amsmath}
\usepackage{booktabs}
\usepackage{xcolor}
\usepackage{subcaption}
\usepackage{arydshln}
\usepackage{tabularx}

\title{Evaluating Multimodal Large Language Models \\ on Vertically Written Japanese Text}

\name{Keito Sasagawa$^{\ast}$$^{\ddagger}$, Shuhei Kurita$^{\dagger}$$^{\ddagger}$, Daisuke Kawahara$^{\ast}$$^{\ddagger}$} 

\address{$^{\ast}$Waseda University, $^{\dagger}$NII, $^{\ddagger}$NII LLMC \\
         Tokyo, Japan \\
         kate@fuji.waseda.jp, skurita@nii.ac.jp, dkw@waseda.jp \\
}

\abstract{
Multimodal Large Language Models (MLLMs) have seen rapid advances in recent years and are now being applied to visual document understanding tasks.
They are expected to process a wide range of document images across languages, including Japanese.
Understanding documents from images requires models to read what are written in them.
Since some Japanese documents are written vertically, support for vertical writing is essential.
However, research specifically focused on vertically written Japanese text remains limited.
In this study, we evaluate the reading capability of existing MLLMs on vertically written Japanese text.
First, we generate a synthetic Japanese OCR dataset by rendering Japanese texts into images, and use it for both model fine-tuning and evaluation.
This dataset includes Japanese text in both horizontal and vertical writing.
We also create an evaluation dataset sourced from the real-world document images containing vertically written Japanese text.
Using these datasets, we demonstrate that the existing MLLMs perform worse on vertically written Japanese text than on horizontally written Japanese text.
Furthermore, we show that training MLLMs on our synthesized Japanese OCR dataset results in improving the performance of models that previously could not handle vertical writing.
The datasets and code are publicly available (\url{https://github.com/llm-jp/eval_vertical_ja}).
 \\ \newline 
 }

\begin{document}

\maketitleabstract

\section{Introduction}

\begin{figure}[t]
  \begin{minipage}[b]{0.95\columnwidth}
    \centering
    \includegraphics[keepaspectratio, width=\columnwidth]{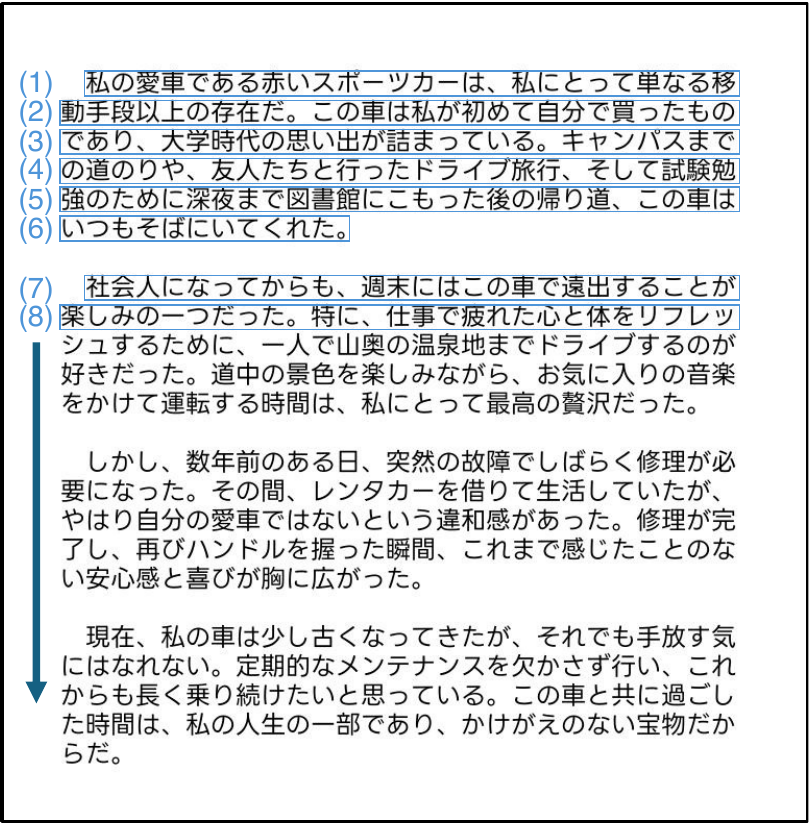}
  \end{minipage} \\
  \begin{minipage}[b]{0.95\columnwidth}
    \centering
    \includegraphics[keepaspectratio, width=\columnwidth]{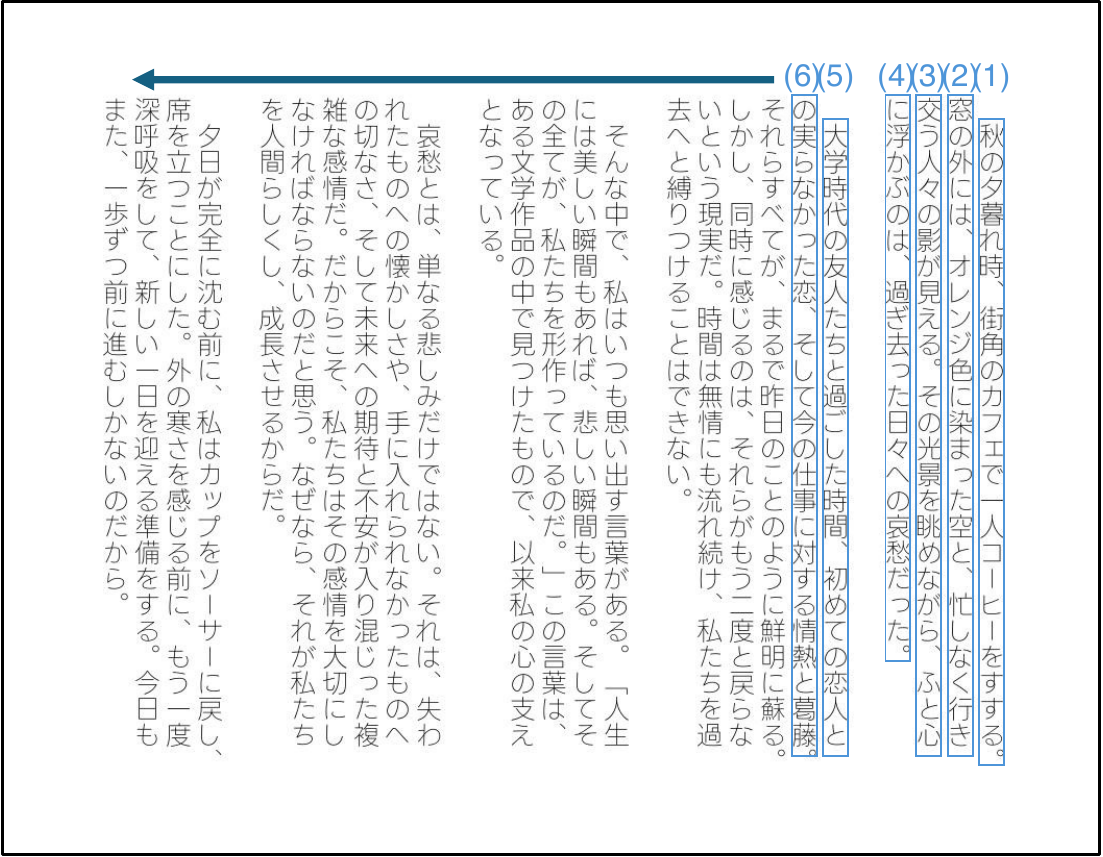}
  \end{minipage}
  \caption{
  Examples of reading order for Japanese documents (Top: Horizontal writing, Bottom: Vertical writing). 
  The \textcolor{blue}{blue numbers} attached to each line indicate the order in which the text on that line should be read.
  In horizontal writing, similar to English documents, characters in each line are read from left to right, and lines are read from top to bottom. 
  In vertical writing, characters in each line are read from top to bottom, and lines are read from right to left.
  }
  \label{fig:example_reading_order}
\end{figure}

Research on multimodal large language models (MLLMs) has advanced rapidly~\cite{bai2025qwen25vltechnicalreport,zhu2025internvl3exploringadvancedtraining,gemmateam2025gemma3technicalreport,openai2025gpt5}, and these models are now applied to a wide range of multimodal tasks.
Document image question answering is one such task, where a model answers questions about given document images, and various datasets have been proposed~\citelanguageresource{Mathew_2021_WACV,Mathew_2022_WACV,Tanaka_Nishida_Yoshida_2021,Tanaka_Nishida_Nishida_Hasegawa_Saito_Saito_2023,TITO2023109834,onami-etal-2024-jdocqa}.
To solve such tasks, MLLMs are required to read the text within the input document image.
A variety of multimodal datasets have been developed for English, and current MLLMs demonstrate high accuracy in reading English text on English-language benchmarks.

Japanese documents often include vertically written text, necessitating explicit support for vertical writing.
First, we explain the difference in reading order between horizontally and vertically written Japanese text.
Figure~\ref{fig:example_reading_order} shows the reading order for horizontally and vertically written Japanese text.
In horizontal Japanese writing, as in English, characters are read from left to right and lines progress from top to bottom.
In vertical writing, characters are read from top to bottom and lines progress from right to left.
It is essential to examine whether models can read vertically written text as well as horizontally written text.
However, most Optical Character Recognition (OCR) benchmarks for MLLMs are primarily concerned with horizontally written text, often neglecting the evaluation of vertically written Japanese text.

To address this gap, we evaluate the OCR capability of MLLMs for reading vertically written Japanese text.
We first build a dataset of synthetic images rendered with Japanese text for training and evaluation.
This dataset comprises images containing text in both horizontal and vertical writing and exhibiting multi-column layouts (1-4 columns).
We also construct an OCR evaluation dataset from real-world PDF pages that contain vertically written Japanese text.

In our experiments, we evaluate multiple open and closed MLLMs, as well as variants fine-tuned on the synthetic image dataset, using our constructed test dataset.
We show that existing MLLMs read vertically written Japanese text less accurately than horizontally written text.
We further demonstrate that fine-tuning on the synthetic dataset improves models that initially struggle with vertically written Japanese text.

The contributions of this study are as follows:
\begin{enumerate}
    \item We evaluate the Japanese OCR capabilities of MLLMs and quantitatively demonstrate that current MLLMs perform worse with vertical writing than with horizontal writing.
    \item We release scripts for synthesizing images of multi-column text in both horizontal and vertical writing styles, along with a dataset of the synthesized images (\textbf{JSSODa}).
    This dataset enables training MLLMs on vertically written text and evaluation.
    \item We also release an OCR dataset containing images of vertically written Japanese text from the real-world PDF pages (\textbf{VJRODa}).
    This dataset enables the evaluation of MLLM's OCR capabilities on real-world document images containing vertically written Japanese text. 
\end{enumerate}
The datasets and code are publicly available (\url{https://github.com/llm-jp/eval_vertical_ja}).

\section{Related Work}
\subsection{Multimodal Large Language Models}

Recent MLLM~\cite{NEURIPS2023_6dcf277e,Liu_2024_CVPR,li2025llavaonevision} architectures employ a structure that connects the vision encoder and LLM via a projection layer.
It converts the input image into features using a vision encoder, feeds them into a projection layer to transform them into image tokens that the LLM can handle, and then inputs them into the LLM together with text tokens.
MLLMs are trained through multimodal instruction tuning, enabling them to handle a wide range of tasks such as document image question answering.

Extensive research has been conducted on multimodal models capable of understanding document images~\cite{10.1145/3394486.3403172,xu-etal-2021-layoutlmv2,10.1145/3503161.3548112,ye2023mplugdocowlmodularizedmultimodallarge,ye-etal-2023-ureader,hu-etal-2024-mplug,hu-etal-2025-mplug,NEURIPS2024_4b06cddd,Luo_2024_CVPR}.
Among these, notable MLLMs that excel at understanding visual Japanese texts and Japanese document images include Qwen2.5-VL~\cite{bai2025qwen25vltechnicalreport}, InternVL3~\cite{zhu2025internvl3exploringadvancedtraining}, and Gemma 3~\cite{gemmateam2025gemma3technicalreport}.
Qwen2.5-VL introduces dynamic resolution processing, enabling native handling of images at various resolutions.
InternVL3 improves performance by employing techniques such as a pixel unshuffle operation, a dynamic resolution strategy that divides images into multiple tiles, and the variable visual position encoding.
Gemma 3 also handles images of various resolutions by dividing the input image into multiple non-overlapping crops of the same size when necessary and feeding them into the vision encoder.

\subsection{Japanese OCR Datasets and Vertical Text Datasets}

NDLOCR~\cite{NDLLabOCR2021} is an OCR program developed by the National Diet Library, Japan (NDL), using materials held in the NDL Digital Collections.
For the evaluation of this program, they used in-house evaluation dataset, which is not publicly available.
Kindai-OCR~\cite{kindaiocr} is an OCR system for modern Japanese magazines based on an attention-based encoder-decoder architecture. 
The document images targeted by this system include vertically written Japanese text, but they are somewhat old and not recent documents.

SVTD and VTD142~\citelanguageresource{10.1007/978-3-030-21074-8_16} are datasets for scene text recognition focusing on vertical text.
SVTD is a synthetic dataset of vertical text images generated following the procedure of~\citet{Cheng_2018_CVPR}, whereas VTD142 is a dataset comprising real-world vertical text collected from web pages.
In~\citetlanguageresource{Oribashi2022}, a synthetic dataset of horizontal and vertical text was constructed based on the synthesis method described in~\citet{Gupta_2016_CVPR} and used for model training.
They also employed the Japanese horizontal and vertical text subsets from the ICDAR 2019 Multi-lingual Scene Text Detection and Recognition competition dataset~\citelanguageresource{8978096} in their experiments.
As the images in these datasets primarily comprise word-level text, sentence-level evaluation is not feasible.

SynthDoG~\citelanguageresource{kim2022donut} is a dataset that comprises images synthesized by compositing document textures onto background images and rendering text.
Although the images in this dataset may partially contain vertical Japanese texts, the proportion is small; moreover, they are unrealistic document images that would not appear in real-world settings, making them unsuitable for an evaluation dataset.
CC-OCR~\citelanguageresource{yang2024ccocrcomprehensivechallengingocr} is a dataset for evaluating MLLM's OCR capabilities and includes Japanese OCR data.
This dataset includes vertical Japanese text, but it is text within natural images and only exists at the word-level, making sentence-level evaluation impossible.
MangaOCR~\citelanguageresource{multimedia_aizawa_2020,mtap_matsui_2017,baek2025mangavqamangalmmbenchmarkspecialized} is an OCR dataset focused on textual elements in manga, such as dialogue and sound effects.
It includes vertical Japanese text; however, its images are restricted to manga pages.

In this work, we construct OCR datasets containing contemporary Japanese text in vertical writing at the sentence-level and evaluate MLLMs.

\section{Dataset Construction}

To evaluate the OCR capabilities of MLLMs on vertically written Japanese text, we construct two types of datasets.
One is a synthetic dataset that can be generated at scale and can also be used for training MLLMs. The other is derived from real-world documents, enabling evaluation under realistic conditions.
These datasets consist of pairs of images and texts written within them.
The first one is \textbf{JSSODa} (\textbf{J}apanese \textbf{S}imple \textbf{S}ynthetic \textbf{O}CR \textbf{Da}taset), which is constructed by rendering Japanese text generated by an LLM into images.
The other one is \textbf{VJRODa} (\textbf{V}ertical \textbf{J}apanese \textbf{R}eal-world \textbf{O}CR \textbf{Da}taset), which consists of images containing vertical Japanese texts sourced from the real-world PDF pages.

\subsection{Construction of JSSODa}

\begin{figure}[t]
  \begin{minipage}[b]{\columnwidth}
    \centering
    \includegraphics[keepaspectratio, width=\columnwidth]{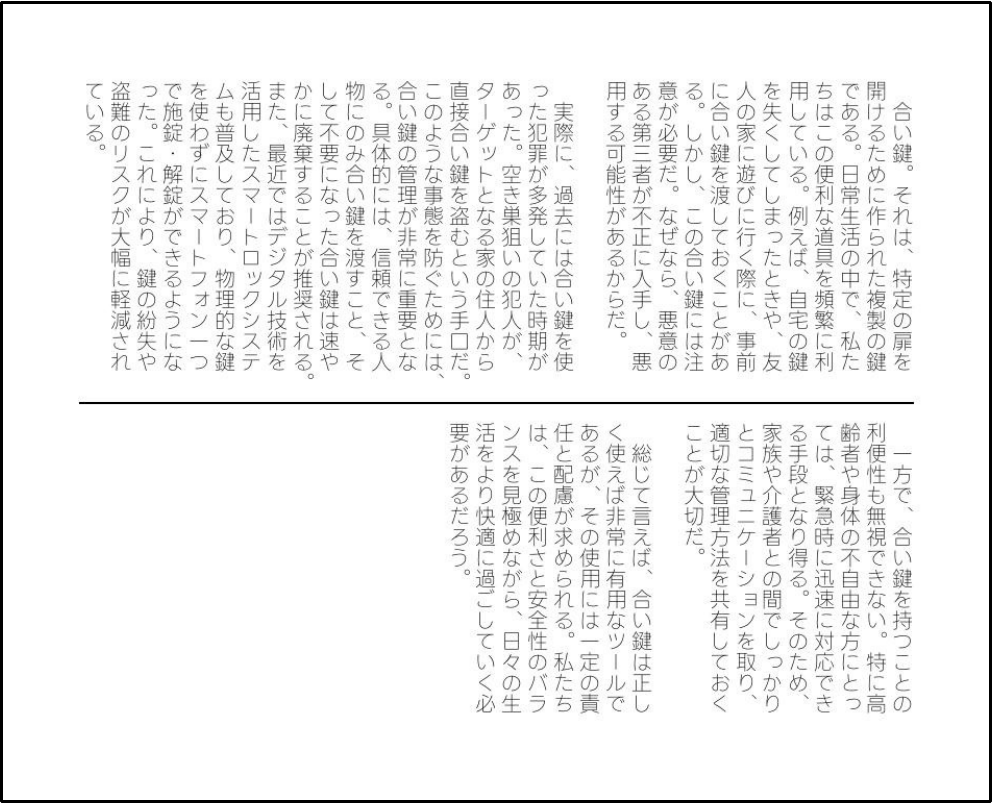}
    \subcaption{vertical, 2-columns}
  \end{minipage} \\
  \begin{minipage}[b]{\columnwidth}
    \centering
    \includegraphics[keepaspectratio, width=\columnwidth]{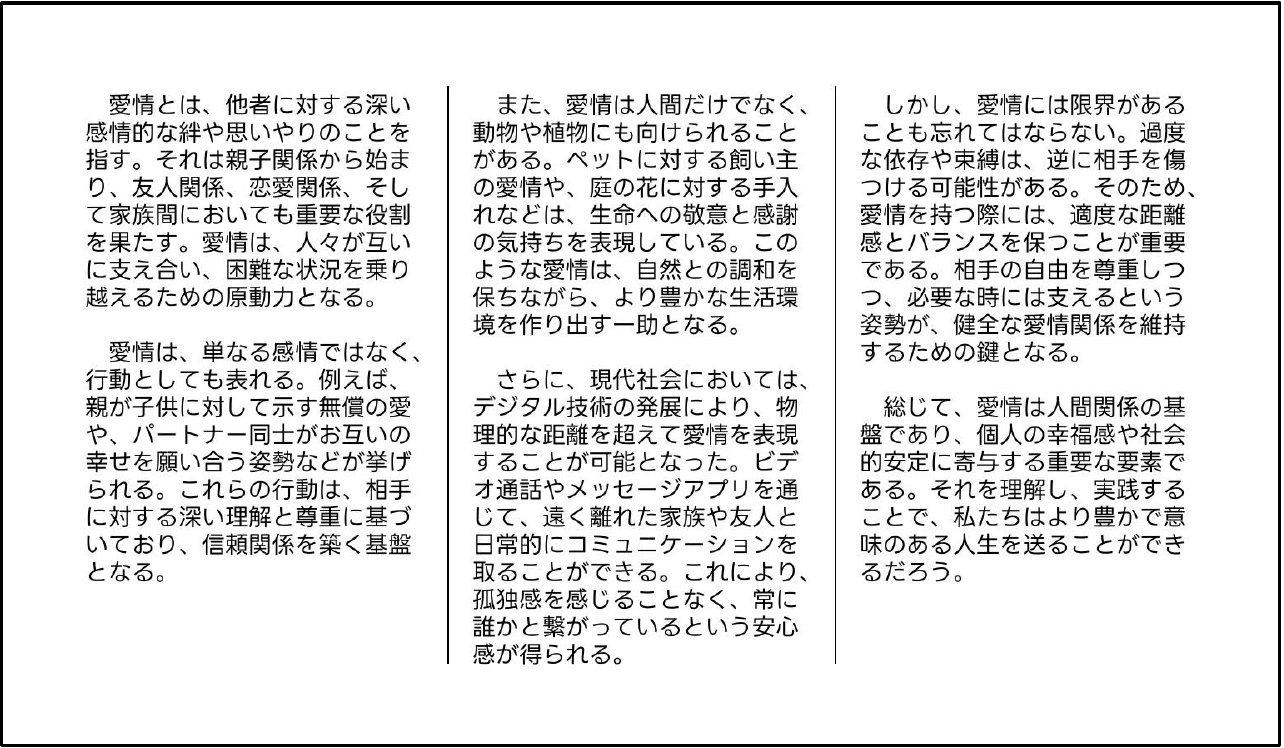}
    \subcaption{horizontal, 3-columns}
  \end{minipage}
  \caption{
  Example images from JSSODa.
  }
  \label{fig:example_image_jssoda}
\end{figure}

To generate large-scale evaluation data, we synthesize images from Japanese texts.
During pre-training, MLLMs are trained using vast amounts of text data, including web text.
If we use web text for our dataset, it may have been used for model training, potentially preventing proper evaluation of the models.
We instead use Japanese text generated by an LLM.

\paragraph{Text Generation}

Specifically, we input Japanese nouns into an LLM and generated sentences about them.
We used nouns from the JUMAN dictionary\footnote{https://github.com/ku-nlp/JumanDIC} for Japanese nouns, and llm-jp-3.1-instruct4~\cite{llmjp} for the LLM.
We removed generated sentences with fewer than 100 characters or more than 3,000 characters.

\paragraph{Image Synthesis}

Next, we synthesized images based on the generated Japanese texts.
We generated a total of 8 layout types of images for both vertical and horizontal writing, each with 1 to 4 columns layout.
Horizontal writing is the same as English text: characters are drawn from left to right, and lines are drawn from top to bottom.
In multi-column layouts, each column is rendered in order from left to right.
In vertical writing, characters are drawn from top to bottom, and lines are drawn from right to left.
In multi-column layouts, each column is rendered in order from top to bottom.
Based on the above character drawing order, we synthesized the images by drawing each character of the generated sentences into images using the Pillow library\footnote{https://github.com/python-pillow/Pillow}.
We collected Japanese fonts from Google Fonts\footnote{https://fonts.google.com/} and free-fonts.jp\footnote{https://free-fonts.jp/}, and used them for synthesizing images.
The number of font files collected is approximately 200.
For the text used, the top 25\% in length was set to four columns, the next 25\%-50\% to three columns, the next 50\%-75\% to two columns, and the top 75\%-100\% to one column.
The total number of images reached 22,493.
Figure~\ref{fig:example_image_jssoda} shows examples of the generated images.

\paragraph{}
We split the constructed dataset into train:val:test at a ratio of 8:1:1 while maintaining the same proportion of the 8 layout types.

\subsection{Construction of VJRODa}

\begin{figure}[t]
    \begin{center}
    \includegraphics[width=\columnwidth]{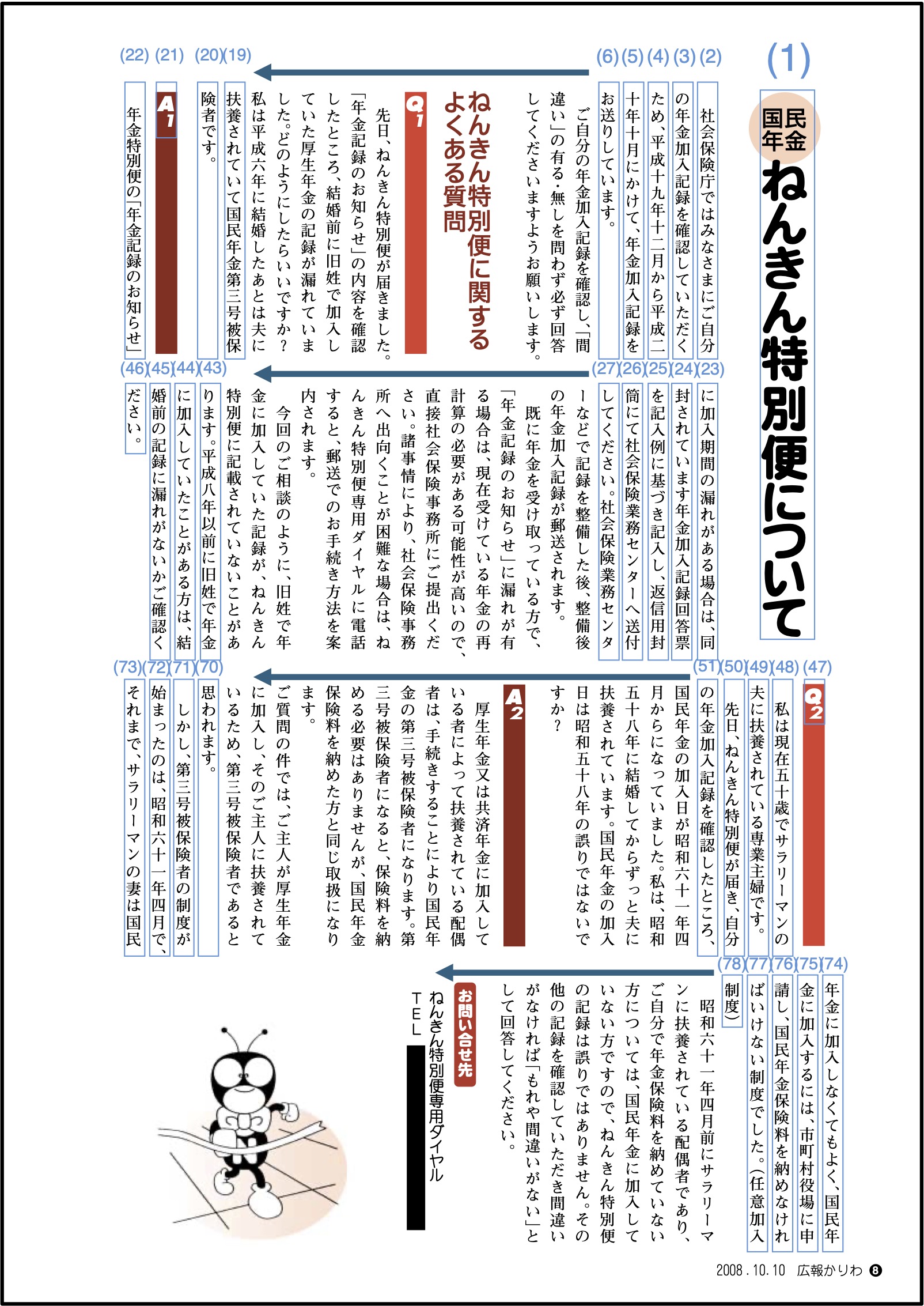}
    \caption{
    An example image from VJRODa.
    The \textcolor{blue}{blue number} above each line indicates the order in which the text on that line should be read. 
    Characters in each line are read from top to bottom, and lines are read from right to left. 
    Each column is read from top to bottom.
    (\url{https://warp.ndl.go.jp/info:ndljp/pid/11712522/www.vill.kariwa.niigata.jp/open/info/000000001_0000000609.pdf}, page 8,
    Personal information has been masked.)
    }
    \label{fig:example_rw_image}
    \end{center}
\end{figure}

We extract pages containing vertically written Japanese text from Japanese PDF documents and annotate each page image with a transcription of its content in Japanese reading order.

First, we collected Japanese PDFs from NDL WARP project\footnote{https://warp.ndl.go.jp/}, and converted each page to an image.
Using these images, we created the dataset following the procedure below.
(1) Filter images that do not contain vertically written Japanese text.
(2) Transcribe the text in the images.

\paragraph{(1) Filter images that do not contain vertically written Japanese text}
We filtered out images that do not contain vertically written Japanese text by using a fast projection profile method~\cite{akiyama1988automatic}, followed by a slower, more detailed vertically written text detection based on character bounding boxes from Tesseract OCR~\cite{smith2007tesseractocr}.
These processes enable the rapid, automatic filtering of images that do not contain vertically written text.

First, each image was converted to grayscale, inverted to obtain black text on a white background, and binarized at a fixed threshold of 128.
We then calculated the total number of binarized 1-valued pixels per row and per column.
For each of them, we calculated the coefficient of variation (CV, standard deviation divided by mean) to examine the degree of variation.
If the column-wise CV is larger, it is treated as a candidate for containing vertically written text.

For images considered to contain vertically written text, we used Tesseract OCR to extract the character-level bounding boxes within them.
We merged the character-level bounding boxes in both the horizontal and vertical directions.
If the number of vertical merges exceeds the number of horizontal merges, we determine that the image contains vertically written text.

Following the filtering steps described above, we manually selected those containing vertically written Japanese text, resulting in a total of 100 images.

\paragraph{(2) Transcribe the text in the images}
We extracted texts from the collected images using PyMuPDF\footnote{https://github.com/pymupdf/PyMuPDF} for pages with embedded text and PaddleOCR~\cite{cui2025paddleocr30technicalreport} for those without, such as scanned images.
Then, we manually corrected the extracted texts to Japanese reading order.
We also corrected OCR errors at the same time.

\paragraph{}
Figure~\ref{fig:example_rw_image} shows an example image in this dataset.

\subsection{Statistics of Datasets}

Table~\ref{table:stats_data} shows the statistics of our datasets.

\begin{table}[t]
\centering
\begin{tabular}{lccc}
\toprule
 & \textbf{JSSODa}  & \textbf{JSSODa} & \textbf{VJRODa} \\
 & \textbf{(train)} & \textbf{(test)} &  \\
\midrule
\# Img              & 17,991  & 2,256  & 100 \\ 
avg. Char           & 706.19  & 705.63 & 1143.91 \\ 
min. Char           & 119     & 192    & 85 \\ 
max. Char           & 1704    & 1399   & 3386 \\ 
\bottomrule
\end{tabular}
\caption{Statistics of our datasets.}
\label{table:stats_data}
\end{table}

\begin{table*}[t]
\centering
\footnotesize
\begin{tabular}{lcccccccc}
\toprule
\multicolumn{9}{c}{\textbf{Horizontal Writing}} \\
\midrule
\textbf{Columns} & \multicolumn{2}{c}{1} & \multicolumn{2}{c}{2} & \multicolumn{2}{c}{3} & \multicolumn{2}{c}{4} \\
\cmidrule(lr){1-1} \cmidrule(lr){2-3} \cmidrule(lr){4-5} \cmidrule(lr){6-7} \cmidrule(lr){8-9}
\textbf{Models} & CER($\downarrow$) & BLEU($\uparrow$) & CER($\downarrow$) & BLEU($\uparrow$) & CER($\downarrow$) & BLEU($\uparrow$) & CER($\downarrow$) & BLEU($\uparrow$) \\
\midrule
\multicolumn{9}{c}{Raw Output} \\
\midrule
Qwen2.5-VL-7B & 7.75 & 91.7 & 19.4 & 87.2 & 21.6 & 87.7 & 23.5 & 82.6 \\
Qwen2.5-VL-32B & \textbf{0.101} & \textbf{99.8} & 17.0 & 99.1 & 10.6 & \textbf{99.2} & 6.79 & \textbf{98.9} \\
InternVL3-8B-hf & 0.462 & 99.3 & 25.3 & 95.4 & 13.6 & 96.3 & 8.20 & 96.9 \\
InternVL3-38B-hf & 0.894 & 98.9 & 10.4 & \textbf{99.3} & 5.15 & 99.0 & 3.78 & 98.0 \\
Gemma 3 12B IT & 3.06 & 95.6 & 16.8 & 92.6 & 15.3 & 92.3 & 17.7 & 84.1 \\
Gemma 3 27B IT & 2.13 & 97.3 & 9.81 & 97.2 & 14.7 & 96.6 & 14.8 & 90.7 \\
\cmidrule(lr){1-1}
GPT-4.1 & 1.88 & 97.8 & \textbf{1.06} & 98.5 & \textbf{1.03} & 98.7 & \textbf{1.10} & 98.4 \\
GPT-5 & 2.09 & 97.5 & 1.26 & 98.3 & 1.28 & 98.6 & 1.96 & 97.4 \\
\midrule
Qwen2.5-VL-7B \scriptsize{(+FT)} & \textbf{0.0637} & \textbf{99.9} & \textbf{0.0251} & \textbf{99.9} & \textbf{0.0266} & \textbf{99.9} & \textbf{0.0383} & \textbf{99.9} \\
InternVL3-8B-hf \scriptsize{(+FT)} & 0.391 & 99.5 & 0.0418 & \textbf{99.9} & 0.0624 & \textbf{99.9} & 0.363 & 99.6 \\
Gemma 3 12B IT \scriptsize{(+FT)} & 0.162 & 99.7 & 0.267 & 99.5 & 0.364 & 99.4 & 1.31 & 98.0 \\
\midrule
\multicolumn{9}{c}{Remove Repetition} \\
\midrule
Qwen2.5-VL-7B & 7.63 & 91.6 & 17.2 & 86.9 & 20.4 & 87.5 & 22.0 & 82.2 \\
Qwen2.5-VL-32B & \textbf{0.101} & \textbf{99.8} & 17.0 & 99.1 & 10.6 & \textbf{99.2} & 6.79 & \textbf{98.9} \\
InternVL3-8B-hf & 0.462 & 99.3 & 25.3 & 95.4 & 13.6 & 96.3 & 7.61 & 96.8 \\
InternVL3-38B-hf & 0.894 & 98.9 & 10.4 & \textbf{99.3} & 5.15 & 99.0 & 3.78 & 98.0 \\
Gemma 3 12B IT & 2.75 & 95.6 & 16.1 & 92.6 & 14.5 & 92.3 & 16.8 & 83.9 \\
Gemma 3 27B IT & 1.66 & 97.4 & 9.81 & 97.2 & 14.7 & 96.6 & 14.8 & 90.7 \\
\cmidrule(lr){1-1}
GPT-4.1 & 1.88 & 97.8 & \textbf{1.06} & 98.5 & \textbf{1.03} & 98.7 & \textbf{1.10} & 98.4 \\
GPT-5 & 2.09 & 97.5 & 1.26 & 98.3 & 1.28 & 98.6 & 1.96 & 97.4 \\
\midrule
Qwen2.5-VL-7B \scriptsize{(+FT)} & \textbf{0.0637} & \textbf{99.9} & \textbf{0.0251} & \textbf{99.9} & \textbf{0.0266} & \textbf{99.9} & \textbf{0.0383} & \textbf{99.9} \\
InternVL3-8B-hf \scriptsize{(+FT)} & 0.391 & 99.5 & 0.0418 & \textbf{99.9} & 0.0624 & \textbf{99.9} & 0.363 & 99.6 \\
Gemma 3 12B IT \scriptsize{(+FT)} & 0.162 & 99.7 & 0.267 & 99.5 & 0.364 & 99.4 & 1.31 & 98.0 \\
\bottomrule
\end{tabular}
\caption{The result on JSSODa test set (horizontal)}
\label{table:eval_jssoda_h}
\end{table*}

\section{Experiments}

\begin{table*}[t]
\centering
\footnotesize
\begin{tabular}{lcccccccc}
\toprule
\multicolumn{9}{c}{\textbf{Vertical Writing}} \\
\midrule
\textbf{Columns} & \multicolumn{2}{c}{1} & \multicolumn{2}{c}{2} & \multicolumn{2}{c}{3} & \multicolumn{2}{c}{4} \\
\cmidrule(lr){1-1} \cmidrule(lr){2-3} \cmidrule(lr){4-5} \cmidrule(lr){6-7} \cmidrule(lr){8-9}
\textbf{Models} & CER($\downarrow$) & BLEU($\uparrow$) & CER($\downarrow$) & BLEU($\uparrow$) & CER($\downarrow$) & BLEU($\uparrow$) & CER($\downarrow$) & BLEU($\uparrow$) \\
\midrule
\multicolumn{9}{c}{Raw Output} \\
\midrule
Qwen2.5-VL-7B & 112 & 26.8 & 100 & 21.1 & 104 & 19.6 & 102 & 18.7 \\
Qwen2.5-VL-32B & 128 & 29.7 & 107 & 18.6 & 104 & 18.7 & 97.9 & 16.6 \\
InternVL3-8B-hf & 81.0 & 50.0 & 64.8 & 61.1 & 64.8 & 61.6 & 61.9 & 59.8 \\
InternVL3-38B-hf & 22.1 & 81.5 & \textbf{43.7} & \textbf{75.4} & 40.1 & 76.5 & 42.7 & \textbf{72.0} \\
Gemma 3 12B IT & 20.3 & 79.2 & 63.3 & 52.8 & 42.8 & 66.4 & 61.1 & 53.0 \\
Gemma 3 27B IT & \textbf{7.62} & \textbf{91.4} & 47.0 & 63.9 & \textbf{28.3} & \textbf{81.3} & \textbf{37.7} & 70.6 \\
\cmidrule(lr){1-1}
GPT-4.1 & 18.2 & 82.8 & 56.0 & 65.8 & 45.9 & 65.2 & 40.7 & 69.4 \\
GPT-5 & 21.3 & 83.1 & 66.0 & 61.7 & 59.4 & 61.6 & 48.7 & 64.7 \\
\midrule
Qwen2.5-VL-7B \scriptsize{(+FT)} & \textbf{0.104} & \textbf{99.8} & \textbf{0.202} & \textbf{99.9} & \textbf{0.113} & \textbf{99.8} & \textbf{0.284} & \textbf{99.6} \\
InternVL3-8B-hf \scriptsize{(+FT)} & 0.619 & 99.4 & 0.315 & 99.8 & 0.330 & 99.6 & 1.10 & 98.8 \\
Gemma 3 12B IT \scriptsize{(+FT)} & 0.502 & 99.1 & 1.04 & 98.5 & 1.47 & 98.1 & 4.09 & 94.8 \\
\midrule
\multicolumn{9}{c}{Remove Repetition} \\
\midrule
Qwen2.5-VL-7B & 73.2 & 31.2 & 83.8 & 19.3 & 88.1 & 17.5 & 87.0 & 15.5 \\
Qwen2.5-VL-32B & 84.2 & 45.7 & 90.6 & 21.0 & 88.6 & 19.3 & 89.6 & 15.9 \\
InternVL3-8B-hf & 39.4 & 66.6 & 47.6 & 68.1 & 50.2 & 70.5 & 49.0 & 67.9 \\
InternVL3-38B-hf & 12.3 & 89.5 & 39.3 & \textbf{74.8} & 34.1 & 80.6 & 35.7 & \textbf{79.3} \\
Gemma 3 12B IT & 14.3 & 83.3 & 48.3 & 57.2 & 35.0 & 69.3 & 52.6 & 54.4 \\
Gemma 3 27B IT & \textbf{7.62} & \textbf{91.4} & \textbf{37.3} & 70.0 & \textbf{26.2} & \textbf{82.6} & \textbf{35.6} & 70.8 \\
\cmidrule(lr){1-1}
GPT-4.1 & 17.4 & 82.7 & 53.8 & 65.5 & 45.5 & 65.1 & 38.6 & 69.1 \\
GPT-5 & 21.3 & 83.1 & 66.0 & 61.7 & 59.4 & 61.6 & 48.7 & 64.7 \\
\midrule
Qwen2.5-VL-7B \scriptsize{(+FT)} & \textbf{0.104} & \textbf{99.8} & \textbf{0.202} & \textbf{99.9} & \textbf{0.113} & \textbf{99.8} & \textbf{0.284} & \textbf{99.6} \\
InternVL3-8B-hf \scriptsize{(+FT)} & 0.619 & 99.4 & 0.315 & 99.8 & 0.330 & 99.6 & 1.10 & 98.8 \\
Gemma 3 12B IT \scriptsize{(+FT)} & 0.502 & 99.1 & 1.04 & 98.5 & 1.47 & 98.1 & 4.09 & 94.8 \\
\bottomrule
\end{tabular}
\caption{The result on JSSODa test set (vertical)}
\label{table:eval_jssoda_v}
\end{table*}

\begin{figure*}[t]
    \begin{center}
    \includegraphics[width=\linewidth]{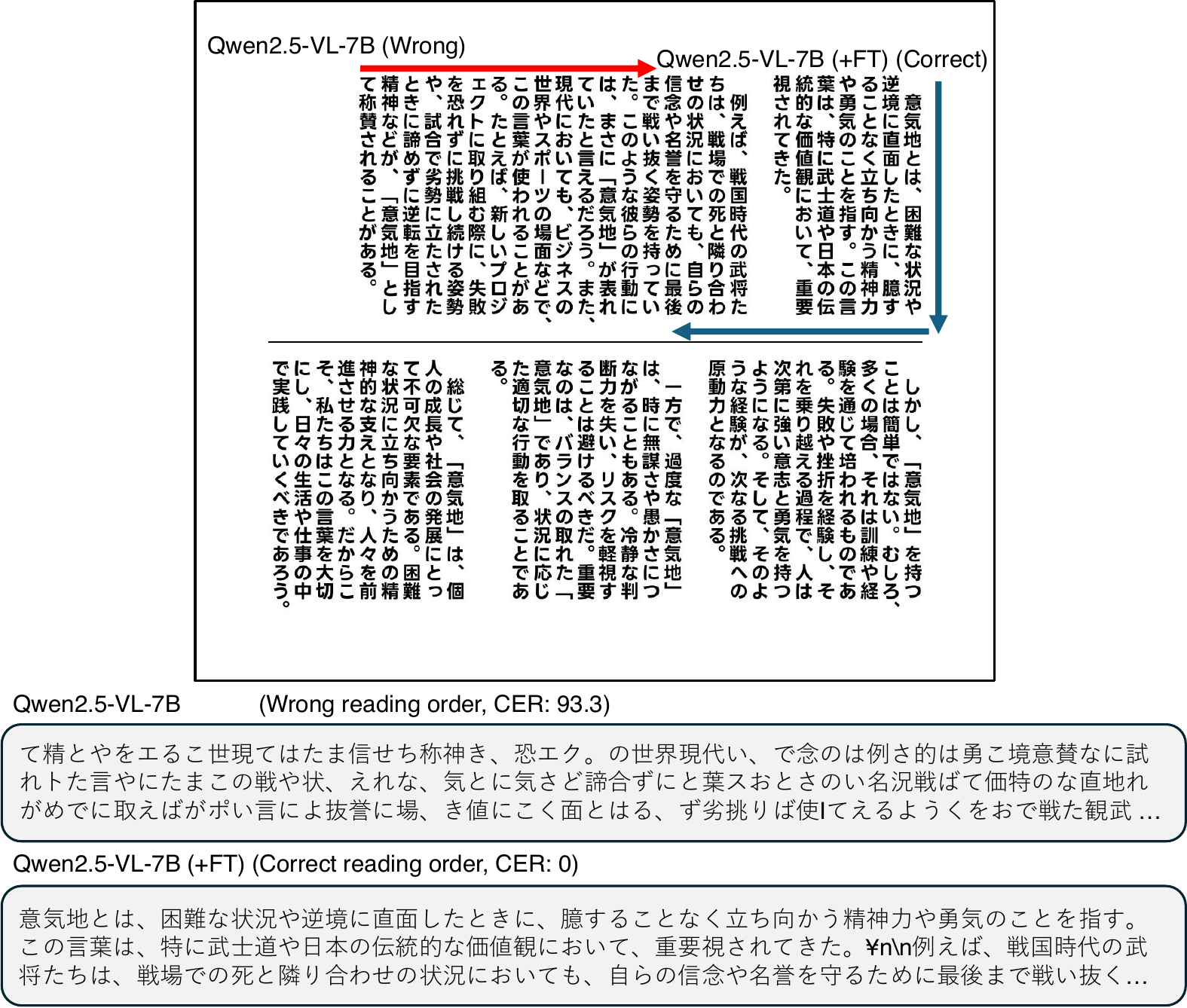}
    \caption{
    The example outputs on JSSODa (vertical) generated by Qwen2.5-VL-7B and its fine-tuned (+FT) variant.
    The \textcolor{red}{red arrow ($\rightarrow$)} represents the reading order of Qwen2.5-VL-7B, 
    and the \textcolor{blue}{blue arrow ($\rightarrow$)} represents the reading order of the fine-tuned model.
    }
    \label{fig:case_study_0}
    \end{center}
\end{figure*}

\subsection{Experimental Setup}
\subsubsection{Fine-tuning Models}
\label{subsubsec:ft_model}

We trained open MLLMs using the JSSODa train set.
As models for training, we used three models capable of reading Japanese text: ``Qwen2.5-VL-7B-Instruct'', ``InternVL3-8B-hf'', and ``Gemma 3 12b IT''.
These models consist of a vision encoder, a multimodal projector, and an LLM.
In this study, we tuned the parameters of all modules.
The number of images used for training was 18k, and we trained the models for one epoch.
We set the batch size to 32.
We used AdamW~\cite{loshchilov2018decoupled} as the optimizer and set the learning rate to 2e-5.

\subsubsection{Evaluation}

We evaluated multiple open MLLMs and closed models on the JSSODa test set and VJRODa.
For open MLLMs, we used Qwen2.5-VL models with 7B and 32B parameters, InternVL3 models with 8B and 38B parameters, and Gemma 3 models with 12B and 27B parameters.
We also evaluated three models fine-tuned on the JSSODa train set, as described in Section~\ref{subsubsec:ft_model}.
During inference, texts were generated using greedy decoding.
As closed models, we used GPT-4.1~\cite{openai2025gpt4-1} and GPT-5~\cite{openai2025gpt5}.
For GPT-4.1, we set the temperature to 0.
For GPT-5, we set ``reasoning\_effort'' to ``minimal''.
For all other parameters, we used the default settings.
We set max new tokens to 1024 for JSSODa test set, and 3072 for VJRODa.
Additionally, we used the same user prompt that was used during the model fine-tuning.

\subsubsection{Evaluation Metric}

For evaluation metrics, we used Character Error Rate (CER) and BLEU~\cite{papineni-etal-2002-bleu}.
Before calculating the scores, we applied Unicode NFKC normalization and whitespace removal to both the reference and predicted texts.

\paragraph{CER} CER is defined by the following formula:
\[
\text{CER} = \frac{\operatorname{EditDistance}(pred, ref)}{|ref|} \times 100,
\]
where $pred$ is the model's predicted text, and $ref$ is the correct text.
$\operatorname{EditDistance}(p, r)$ is the edit distance between strings $p$ and $r$.
Lower CER values are generally considered indicative of better performance.

\paragraph{BLEU}

We used SacreBLEU~\cite{post-2018-call} for calculating BLEU.
The reference texts and predicted texts were split into character units.

\paragraph{}

MLLMs sometimes generate the same tokens repeatedly, up to max new tokens.
When this behavior happens, the scores tend to become extremely low and may not be reliable as a reference for evaluation.
To cope with this problem, we also report the scores when repetitive sections are removed from the predicted texts.
We use a regular expression to remove the last ten or more consecutive occurrences of a string from the entire string, leaving only the first occurrence.

\subsection{Results}
\subsubsection{Result on JSSODa}

Tables~\ref{table:eval_jssoda_h} and~\ref{table:eval_jssoda_v} show the evaluation results on the JSSODa test set.
Each shows the results for horizontal and vertical writing.
The part below ``Raw Output'' shows the results when MLLM's output was used directly for evaluation, while the part below ``Remove Repetition'' shows the results after removing repeated sections.
``(+FT)'' denotes the model fine-tuned on the JSSODa train set.

The results highlight a clear disparity: while all models handled horizontally written text reasonably well, they struggled significantly with vertically written text.
In particular, Qwen2.5-VL sometimes read vertically written text in horizontal reading order, as shown in Section~\ref{subsec:case_study}.
InternVL3, Gemma 3, GPT-4.1, and GPT-5 seemed to have a certain understanding of character reading order, but compared to horizontally written text, they made more errors in character recognition.
For both InternVL3 and Gemma 3, we observe that performance on vertically written text tends to improve as the model's parameter count increases.
GPT-5 sometimes produced no text output at all.
The reasoning trace alone may have reached the max new tokens limit.
Additionally, removing repetitions improves the score, indicating that MLLMs generate tokens repeatedly.
When the model is trained on the train set, the score for vertically written text improves substantially.

\subsubsection{Result on VJRODa}

\begin{table*}[t]
\centering
\footnotesize
\begin{tabular}{lcccc}
\toprule
 & \multicolumn{2}{c}{Raw Output} & \multicolumn{2}{c}{Remove Repetition} \\
 \cmidrule(lr){2-3} \cmidrule(lr){4-5}
\textbf{Models} & CER($\downarrow$) & BLEU($\uparrow$) & CER($\downarrow$) & BLEU($\uparrow$) \\
\midrule
Qwen2.5-VL-7B              & 154  & 20.1 & 88.5 & 22.0  \\ 
Qwen2.5-VL-32B             & 128  & \textbf{42.6} & 63.0 & \textbf{58.5} \\ 
InternVL3-8B-hf            & 121  & 26.0 & 66.5 & 40.8 \\ 
InternVL3-38B-hf           & 138  & 27.7 & 64.0 & 45.1 \\ 
Gemma 3 12B IT             & 125  & 17.5 & 67.9 & 23.3 \\ 
Gemma 3 27B IT             & \textbf{67.9} & 35.0 & \textbf{56.7} & 34.2 \\ 
\cmidrule(lr){1-1}
GPT-4.1                    & 101  & 29.2 & 61.7 & 34.1 \\ 
GPT-5                      & 70.1 & 40.9 & 69.4 & 41.0 \\ 
\midrule
Qwen2.5-VL-7B \scriptsize{(+FT)}       & \textbf{65.1} & \textbf{51.5} & \textbf{40.5} & \textbf{61.1} \\ 
InternVL3-8B-hf \scriptsize{(+FT)}     & 251 & 26.1 & 73.5 & 54.9 \\
Gemma 3 12B IT \scriptsize{(+FT)}      & 77.6 & 27.9 & 67.4 & 27.2 \\
\bottomrule
\end{tabular}
\caption{The result on VJRODa}
\label{table:eval_rw}
\end{table*}

Table~\ref{table:eval_rw} shows the evaluation result on VJRODa.
None of the models performed well, indicating that they struggle with text in real-world vertical-writing document images.
When fine-tuned on the JSSODa train set, Qwen2.5-VL-7B exhibited substantial performance gains under both the ``Raw Output'' and ``Remove Repetition'' evaluation settings, whereas the other two models showed no appreciable improvement.
A plausible explanation is that Qwen2.5-VL-7B initially lacked a robust understanding of the reading order in vertically written Japanese text, while the other models had already internalized this to some extent.
These findings suggest that the JSSODa train set may be effective for enhancing vertical-text reading capabilities in models that do not yet capture Japanese vertical reading order.
By contrast, improving performance on vertically written text in real-world documents likely requires training with real-world OCR datasets.

\subsection{Case Study}
\label{subsec:case_study}

\begin{figure*}[t]
    \begin{center}
    \includegraphics[width=\linewidth]{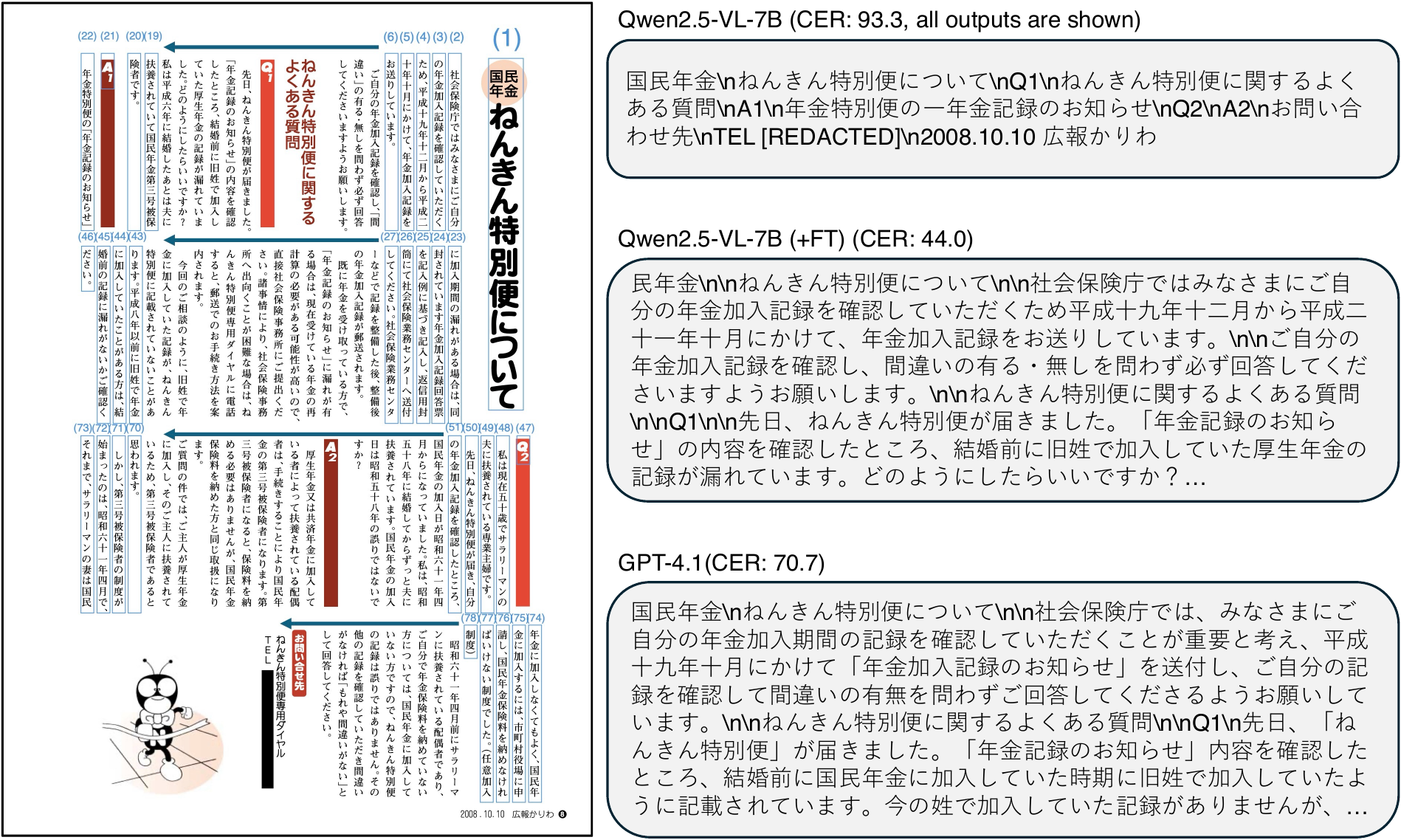}
    \caption{
    The example outputs on VJRODa generated by Qwen2.5-VL-7B, its fine-tuned (+FT) variant, and GPT-4.1.
    (\url{https://warp.ndl.go.jp/info:ndljp/pid/11712522/www.vill.kariwa.niigata.jp/open/info/000000001_0000000609.pdf}, page 8,
    Personal information has been masked.)
    }
    \label{fig:case_study_1}
    \end{center}
\end{figure*}

Figure~\ref{fig:case_study_0} shows example outputs for images from the JSSODa test set (vertical).
The original Qwen2.5-VL-7B incorrectly read vertically written text from top-left to right.
In the fine-tuned model, vertically written text could now be read correctly, in order from the top right downward.

Figure~\ref{fig:case_study_1} shows example outputs from real-world data (VJRODa).
Qwen2.5-VL-7B was only able to output a portion of the text.
On the other hand, the fine-tuned model output text mostly following Japanese reading order, though with some errors.
In GPT-4.1, the output appears to adhere to the reading order of vertically written Japanese text; however, some portions of the output do not correspond to the actual text present in the image.

\section{Conclusion}

In this paper, we evaluated the OCR capability of MLLMs for vertically written Japanese text.
To this end, we constructed a synthetic image dataset of horizontally and vertically written Japanese text, as well as a dataset of document images containing vertically written Japanese text collected from real-world PDFs, and conducted evaluations.
Evaluation results indicate that the current MLLMs struggle with reading vertically written Japanese text.
We also found that training on our dataset of synthetic images can improve the performance of a model that does not handle vertically written Japanese text well.
In the future, we would like to explore methods for building models that can handle a variety of Japanese document images.

\section{Acknowledgement}

In this work, we used the ``mdx: a platform for building data-empowered society''~\cite{9927975mdx}.

\section{Limitations}

\paragraph{The Need for a More Visually Diverse Synthetic Japanese OCR Dataset for Model Training}

Our JSSODa dataset comprises clean, minimalist document images consisting of black text on a white background; no synthetic noise or other degradations are introduced, and no embedded images, tables, or figures are included.
To better generalize to real-world document images, it is necessary to develop a dataset that encompasses greater visual diversity.

\paragraph{Evaluation on Documents with Multiple Plausible Reading Orders}

In real-world document images, there can be multiple plausible reading orders.
For instance, the text within an image caption may have several possible reading sequences.
In our VJRODa dataset, only a single reading order is annotated for each document. Consequently, under metrics such as CER and BLEU, even a correct reading order can potentially receive a low score if it differs from the single ground truth.
Ideally, one would annotate all possible ground-truth reading orders and select the one that yields the highest score; however, this approach is very expensive.
Therefore, developing an efficient and effective method for evaluating the OCR capabilities of MLLMs is an important direction for future work.

\paragraph{Image Diversity in the VJRODa Dataset}

The images in the VJRODa dataset are sourced primarily from PDFs obtained from public-sector websites.
As a result, the domain of document images is relatively narrow and may lack visual diversity.

\section{Ethical Considerations}

Since our JSSODa dataset is constructed from LLM-generated texts, it does not infringe third-party copyrights.
Regarding the data sources of our VJRODa dataset, their use for information analysis is permitted under the copyright law of the country in which the research was conducted.

\nocite{*}
\section{Bibliographical References}\label{sec:reference}

\bibliographystyle{lrec2026-natbib}
\bibliography{lrec2026-example}

\begin{thebibliography}{14}
\expandafter\ifx\csname natexlab\endcsname\relax\def\natexlab#1{#1}\fi

\bibitem[{Aizawa et~al.(2020)Aizawa, Fujimoto, Otsubo, Ogawa, Matsui, Tsubota, and Ikuta}]{multimedia_aizawa_2020}
Kiyoharu Aizawa, Azuma Fujimoto, Atsushi Otsubo, Toru Ogawa, Yusuke Matsui, Koki Tsubota, and Hikaru Ikuta. 2020.
\newblock \href {https://doi.org/10.1109/mmul.2020.2987895} {Building a manga dataset ``manga109'' with annotations for multimedia applications}.
\newblock \emph{IEEE MultiMedia}, 27(2):8--18.

\bibitem[{Baek et~al.(2025)Baek, Egashira, Onohara, Miyai, Imajuku, Ikuta, and Aizawa}]{baek2025mangavqamangalmmbenchmarkspecialized}
Jeonghun Baek, Kazuki Egashira, Shota Onohara, Atsuyuki Miyai, Yuki Imajuku, Hikaru Ikuta, and Kiyoharu Aizawa. 2025.
\newblock \href {http://arxiv.org/abs/2505.20298} {Mangavqa and mangalmm: A benchmark and specialized model for multimodal manga understanding}.

\bibitem[{Choi et~al.(2019)Choi, Yoon, Lee, and Kim}]{10.1007/978-3-030-21074-8_16}
Chankyu Choi, Youngmin Yoon, Junsu Lee, and Junseok Kim. 2019.
\newblock Simultaneous recognition of horizontal and vertical text in natural images.
\newblock In \emph{Computer Vision -- ACCV 2018 Workshops}, pages 202--212, Cham. Springer International Publishing.

\bibitem[{Kim et~al.(2022)Kim, Hong, Yim, Nam, Park, Yim, Hwang, Yun, Han, and Park}]{kim2022donut}
Geewook Kim, Teakgyu Hong, Moonbin Yim, JeongYeon Nam, Jinyoung Park, Jinyeong Yim, Wonseok Hwang, Sangdoo Yun, Dongyoon Han, and Seunghyun Park. 2022.
\newblock Ocr-free document understanding transformer.
\newblock In \emph{European Conference on Computer Vision (ECCV)}.

\bibitem[{Mathew et~al.(2022)Mathew, Bagal, Tito, Karatzas, Valveny, and Jawahar}]{Mathew_2022_WACV}
Minesh Mathew, Viraj Bagal, Rub\`en Tito, Dimosthenis Karatzas, Ernest Valveny, and C.V. Jawahar. 2022.
\newblock Infographicvqa.
\newblock In \emph{Proceedings of the IEEE/CVF Winter Conference on Applications of Computer Vision (WACV)}, pages 1697--1706.

\bibitem[{Mathew et~al.(2021)Mathew, Karatzas, and Jawahar}]{Mathew_2021_WACV}
Minesh Mathew, Dimosthenis Karatzas, and C.V. Jawahar. 2021.
\newblock Docvqa: A dataset for vqa on document images.
\newblock In \emph{Proceedings of the IEEE/CVF Winter Conference on Applications of Computer Vision (WACV)}, pages 2200--2209.

\bibitem[{Matsui et~al.(2017)Matsui, Ito, Aramaki, Fujimoto, Ogawa, Yamasaki, and Aizawa}]{mtap_matsui_2017}
Yusuke Matsui, Kota Ito, Yuji Aramaki, Azuma Fujimoto, Toru Ogawa, Toshihiko Yamasaki, and Kiyoharu Aizawa. 2017.
\newblock \href {https://doi.org/10.1007/s11042-016-4020-z} {Sketch-based manga retrieval using manga109 dataset}.
\newblock \emph{Multimedia Tools and Applications}, 76(20):21811--21838.

\bibitem[{Nayef et~al.(2019)Nayef, Patel, Busta, Chowdhury, Karatzas, Khlif, Matas, Pal, Burie, Liu, and Ogier}]{8978096}
Nibal Nayef, Yash Patel, Michal Busta, Pinaki~Nath Chowdhury, Dimosthenis Karatzas, Wafa Khlif, Jiri Matas, Umapada Pal, Jean-Christophe Burie, Cheng-lin Liu, and Jean-Marc Ogier. 2019.
\newblock \href {https://doi.org/10.1109/ICDAR.2019.00254} {Icdar2019 robust reading challenge on multi-lingual scene text detection and recognition — rrc-mlt-2019}.
\newblock In \emph{2019 International Conference on Document Analysis and Recognition (ICDAR)}, pages 1582--1587.

\bibitem[{Onami et~al.(2024)Onami, Kurita, Miyanishi, and Watanabe}]{onami-etal-2024-jdocqa}
Eri Onami, Shuhei Kurita, Taiki Miyanishi, and Taro Watanabe. 2024.
\newblock \href {https://aclanthology.org/2024.lrec-main.830/} {{JD}oc{QA}: {J}apanese document question answering dataset for generative language models}.
\newblock In \emph{Proceedings of the 2024 Joint International Conference on Computational Linguistics, Language Resources and Evaluation (LREC-COLING 2024)}, pages 9503--9514, Torino, Italia. ELRA and ICCL.

\bibitem[{Orihashi et~al.(2022)Orihashi, Yamazaki, Uchida, Takashima, and Masumura}]{Oribashi2022}
Shota Orihashi, Yoshihiro Yamazaki, Mihiro Uchida, Akihiko Takashima, and Ryo Masumura. 2022.
\newblock \href {https://www.ieice.org/publications/conference-FIT-DVDs/FIT2022/data/html/program/pdf/CH-012.pdf} {Shared modeling of horizontal and vertical writing using character counting for japanese scene text recognition}.
\newblock In \emph{Proceedings of the 21st Forum on Information Technology (FIT2022)}.

\bibitem[{Tanaka et~al.(2023)Tanaka, Nishida, Nishida, Hasegawa, Saito, and Saito}]{Tanaka_Nishida_Nishida_Hasegawa_Saito_Saito_2023}
Ryota Tanaka, Kyosuke Nishida, Kosuke Nishida, Taku Hasegawa, Itsumi Saito, and Kuniko Saito. 2023.
\newblock \href {https://doi.org/10.1609/aaai.v37i11.26598} {Slidevqa: A dataset for document visual question answering on multiple images}.
\newblock \emph{Proceedings of the AAAI Conference on Artificial Intelligence}, 37(11):13636--13645.

\bibitem[{Tanaka et~al.(2021)Tanaka, Nishida, and Yoshida}]{Tanaka_Nishida_Yoshida_2021}
Ryota Tanaka, Kyosuke Nishida, and Sen Yoshida. 2021.
\newblock \href {https://doi.org/10.1609/aaai.v35i15.17635} {Visualmrc: Machine reading comprehension on document images}.
\newblock \emph{Proceedings of the AAAI Conference on Artificial Intelligence}, 35(15):13878--13888.

\bibitem[{Tito et~al.(2023)Tito, Karatzas, and Valveny}]{TITO2023109834}
Rubèn Tito, Dimosthenis Karatzas, and Ernest Valveny. 2023.
\newblock \href {https://doi.org/https://doi.org/10.1016/j.patcog.2023.109834} {Hierarchical multimodal transformers for multipage docvqa}.
\newblock \emph{Pattern Recognition}, 144:109834.

\bibitem[{Yang et~al.(2024)Yang, Tang, Li, Wang, Wan, Zhong, Liu, Yang, Wang, Bai, Jin, and Lin}]{yang2024ccocrcomprehensivechallengingocr}
Zhibo Yang, Jun Tang, Zhaohai Li, Pengfei Wang, Jianqiang Wan, Humen Zhong, Xuejing Liu, Mingkun Yang, Peng Wang, Shuai Bai, LianWen Jin, and Junyang Lin. 2024.
\newblock \href {http://arxiv.org/abs/2412.02210} {Cc-ocr: A comprehensive and challenging ocr benchmark for evaluating large multimodal models in literacy}.

\end{thebibliography}


\begin{thebibliography}{31}
\expandafter\ifx\csname natexlab\endcsname\relax\def\natexlab#1{#1}\fi

\bibitem[{Akiyama and Hagita(1988)}]{akiyama1988automatic}
Teruo Akiyama and Norihiro Hagita. 1988.
\newblock Automatic reading system for printed documents.
\newblock In \emph{Proceedings of IAPR Workshop on COMPUTER VISION}.

\bibitem[{Bai et~al.(2025)Bai, Chen, Liu, Wang, Ge, Song, Dang, Wang, Wang, Tang, Zhong, Zhu, Yang, Li, Wan, Wang, Ding, Fu, Xu, Ye, Zhang, Xie, Cheng, Zhang, Yang, Xu, and Lin}]{bai2025qwen25vltechnicalreport}
Shuai Bai, Keqin Chen, Xuejing Liu, Jialin Wang, Wenbin Ge, Sibo Song, Kai Dang, Peng Wang, Shijie Wang, Jun Tang, Humen Zhong, Yuanzhi Zhu, Mingkun Yang, Zhaohai Li, Jianqiang Wan, Pengfei Wang, Wei Ding, Zheren Fu, Yiheng Xu, Jiabo Ye, Xi~Zhang, Tianbao Xie, Zesen Cheng, Hang Zhang, Zhibo Yang, Haiyang Xu, and Junyang Lin. 2025.
\newblock \href {http://arxiv.org/abs/2502.13923} {Qwen2.5-vl technical report}.

\bibitem[{Cheng et~al.(2018)Cheng, Xu, Bai, Niu, Pu, and Zhou}]{Cheng_2018_CVPR}
Zhanzhan Cheng, Yangliu Xu, Fan Bai, Yi~Niu, Shiliang Pu, and Shuigeng Zhou. 2018.
\newblock Aon: Towards arbitrarily-oriented text recognition.
\newblock In \emph{Proceedings of the IEEE Conference on Computer Vision and Pattern Recognition (CVPR)}.

\bibitem[{Cui et~al.(2025{\natexlab{a}})Cui, Sun, Liang, Gao, Zhang, Liu, Wang, Zhou, Liu, Lin, Zhang, Zhang, Zheng, Zhang, Zhang, Liu, Yu, and Ma}]{cui2025paddleocrvlboostingmultilingualdocument}
Cheng Cui, Ting Sun, Suyin Liang, Tingquan Gao, Zelun Zhang, Jiaxuan Liu, Xueqing Wang, Changda Zhou, Hongen Liu, Manhui Lin, Yue Zhang, Yubo Zhang, Handong Zheng, Jing Zhang, Jun Zhang, Yi~Liu, Dianhai Yu, and Yanjun Ma. 2025{\natexlab{a}}.
\newblock \href {http://arxiv.org/abs/2510.14528} {Paddleocr-vl: Boosting multilingual document parsing via a 0.9b ultra-compact vision-language model}.

\bibitem[{Cui et~al.(2025{\natexlab{b}})Cui, Sun, Lin, Gao, Zhang, Liu, Wang, Zhang, Zhou, Liu, Zhang, Lv, Huang, Zhang, Zhang, Zhang, Liu, Yu, and Ma}]{cui2025paddleocr30technicalreport}
Cheng Cui, Ting Sun, Manhui Lin, Tingquan Gao, Yubo Zhang, Jiaxuan Liu, Xueqing Wang, Zelun Zhang, Changda Zhou, Hongen Liu, Yue Zhang, Wenyu Lv, Kui Huang, Yichao Zhang, Jing Zhang, Jun Zhang, Yi~Liu, Dianhai Yu, and Yanjun Ma. 2025{\natexlab{b}}.
\newblock \href {http://arxiv.org/abs/2507.05595} {Paddleocr 3.0 technical report}.

\bibitem[{Dong et~al.(2024)Dong, Zhang, Zang, Cao, Wang, Ouyang, Zhang, Duan, Zhang, Li, Yan, Gao, Chen, Zhang, Li, Li, Wang, Chen, He, Zhang, Dai, Qiao, Lin, and Wang}]{NEURIPS2024_4b06cddd}
Xiaoyi Dong, Pan Zhang, Yuhang Zang, Yuhang Cao, Bin Wang, Linke Ouyang, Songyang Zhang, Haodong Duan, Wenwei Zhang, Yining Li, Hang Yan, Yang Gao, Zhe Chen, Xinyue Zhang, Wei Li, Jingwen Li, Wenhai Wang, Kai Chen, Conghui He, Xingcheng Zhang, Jifeng Dai, Yu~Qiao, Dahua Lin, and Jiaqi Wang. 2024.
\newblock \href {https://proceedings.neurips.cc/paper_files/paper/2024/file/4b06cdddb1cde6624c0be1465c7b800f-Paper-Conference.pdf} {Internlm-xcomposer2-4khd: A pioneering large vision-language model handling resolutions from 336 pixels to 4k hd}.
\newblock In \emph{Advances in Neural Information Processing Systems}, volume~37, pages 42566--42592. Curran Associates, Inc.

\bibitem[{Gupta et~al.(2016)Gupta, Vedaldi, and Zisserman}]{Gupta_2016_CVPR}
Ankush Gupta, Andrea Vedaldi, and Andrew Zisserman. 2016.
\newblock Synthetic data for text localisation in natural images.
\newblock In \emph{Proceedings of the IEEE Conference on Computer Vision and Pattern Recognition (CVPR)}.

\bibitem[{Hu et~al.(2024)Hu, Xu, Ye, Yan, Zhang, Zhang, Zhang, Jin, Huang, and Zhou}]{hu-etal-2024-mplug}
Anwen Hu, Haiyang Xu, Jiabo Ye, Ming Yan, Liang Zhang, Bo~Zhang, Ji~Zhang, Qin Jin, Fei Huang, and Jingren Zhou. 2024.
\newblock \href {https://doi.org/10.18653/v1/2024.findings-emnlp.175} {m{PLUG}-{D}oc{O}wl 1.5: Unified structure learning for {OCR}-free document understanding}.
\newblock In \emph{Findings of the Association for Computational Linguistics: EMNLP 2024}, pages 3096--3120, Miami, Florida, USA. Association for Computational Linguistics.

\bibitem[{Hu et~al.(2025)Hu, Xu, Zhang, Ye, Yan, Zhang, Jin, Huang, and Zhou}]{hu-etal-2025-mplug}
Anwen Hu, Haiyang Xu, Liang Zhang, Jiabo Ye, Ming Yan, Ji~Zhang, Qin Jin, Fei Huang, and Jingren Zhou. 2025.
\newblock \href {https://doi.org/10.18653/v1/2025.acl-long.291} {m{PLUG}-{D}oc{O}wl2: High-resolution compressing for {OCR}-free multi-page document understanding}.
\newblock In \emph{Proceedings of the 63rd Annual Meeting of the Association for Computational Linguistics (Volume 1: Long Papers)}, pages 5817--5834, Vienna, Austria. Association for Computational Linguistics.

\bibitem[{Huang et~al.(2022)Huang, Lv, Cui, Lu, and Wei}]{10.1145/3503161.3548112}
Yupan Huang, Tengchao Lv, Lei Cui, Yutong Lu, and Furu Wei. 2022.
\newblock \href {https://doi.org/10.1145/3503161.3548112} {Layoutlmv3: Pre-training for document ai with unified text and image masking}.
\newblock In \emph{Proceedings of the 30th ACM International Conference on Multimedia}, MM '22, page 4083–4091, New York, NY, USA. Association for Computing Machinery.

\bibitem[{Le et~al.(2019)Le, Mochihashi, Masuda, Mima, and Ly}]{kindaiocr}
Anh~Duc Le, Daichi Mochihashi, Katsuya Masuda, Hideki Mima, and Nam~Tuan Ly. 2019.
\newblock \href {https://doi.org/10.1145/3352631.3352641} {Recognition of japanese historical text lines by an attention-based encoder-decoder and text line generation}.
\newblock In \emph{Proceedings of the 5th International Workshop on Historical Document Imaging and Processing}, HIP '19, page 37–41, New York, NY, USA. Association for Computing Machinery.

\bibitem[{Li et~al.(2025)Li, Zhang, Guo, Zhang, Li, Zhang, Zhang, Zhang, Li, Liu, and Li}]{li2025llavaonevision}
Bo~Li, Yuanhan Zhang, Dong Guo, Renrui Zhang, Feng Li, Hao Zhang, Kaichen Zhang, Peiyuan Zhang, Yanwei Li, Ziwei Liu, and Chunyuan Li. 2025.
\newblock \href {https://openreview.net/forum?id=zKv8qULV6n} {{LL}a{VA}-onevision: Easy visual task transfer}.
\newblock \emph{Transactions on Machine Learning Research}.

\bibitem[{Liu et~al.(2024)Liu, Li, Li, and Lee}]{Liu_2024_CVPR}
Haotian Liu, Chunyuan Li, Yuheng Li, and Yong~Jae Lee. 2024.
\newblock Improved baselines with visual instruction tuning.
\newblock In \emph{Proceedings of the IEEE/CVF Conference on Computer Vision and Pattern Recognition (CVPR)}, pages 26296--26306.

\bibitem[{Liu et~al.(2023)Liu, Li, Wu, and Lee}]{NEURIPS2023_6dcf277e}
Haotian Liu, Chunyuan Li, Qingyang Wu, and Yong~Jae Lee. 2023.
\newblock \href {https://proceedings.neurips.cc/paper_files/paper/2023/file/6dcf277ea32ce3288914faf369fe6de0-Paper-Conference.pdf} {Visual instruction tuning}.
\newblock In \emph{Advances in Neural Information Processing Systems}, volume~36, pages 34892--34916. Curran Associates, Inc.

\bibitem[{LLM-jp(2024)}]{llmjp}
LLM-jp. 2024.
\newblock \href {https://doi.org/10.48550/arXiv.2407.03963} {Llm-jp: A cross-organizational project for the research and development of fully open japanese llms}.
\newblock \emph{CoRR}, abs/2407.03963.

\bibitem[{Loshchilov and Hutter(2019)}]{loshchilov2018decoupled}
Ilya Loshchilov and Frank Hutter. 2019.
\newblock \href {https://openreview.net/forum?id=Bkg6RiCqY7} {Decoupled weight decay regularization}.
\newblock In \emph{International Conference on Learning Representations}.

\bibitem[{Luo et~al.(2024)Luo, Shen, Zhu, Zheng, Yu, and Yao}]{Luo_2024_CVPR}
Chuwei Luo, Yufan Shen, Zhaoqing Zhu, Qi~Zheng, Zhi Yu, and Cong Yao. 2024.
\newblock Layoutllm: Layout instruction tuning with large language models for document understanding.
\newblock In \emph{Proceedings of the IEEE/CVF Conference on Computer Vision and Pattern Recognition (CVPR)}, pages 15630--15640.

\bibitem[{{NDL Lab}(2021)}]{NDLLabOCR2021}
{NDL Lab}. 2021.
\newblock Development of japanese ocr software in fy2021.
\newblock \url{https://lab.ndl.go.jp/data_set/ocr_en/r3_software/}.
\newblock Accessed: 2025-09-24.

\bibitem[{OpenAI(2025{\natexlab{a}})}]{openai2025gpt4-1}
OpenAI. 2025{\natexlab{a}}.
\newblock Introducing gpt-4.1 in the api.
\newblock \url{https://openai.com/index/gpt-4-1/}.
\newblock Accessed: 2025-09-24.

\bibitem[{OpenAI(2025{\natexlab{b}})}]{openai2025gpt5}
OpenAI. 2025{\natexlab{b}}.
\newblock Introducing gpt-5.
\newblock \url{https://openai.com/index/introducing-gpt-5/}.
\newblock Accessed: 2025-09-24.

\bibitem[{Papineni et~al.(2002)Papineni, Roukos, Ward, and Zhu}]{papineni-etal-2002-bleu}
Kishore Papineni, Salim Roukos, Todd Ward, and Wei-Jing Zhu. 2002.
\newblock \href {https://doi.org/10.3115/1073083.1073135} {{B}leu: a method for automatic evaluation of machine translation}.
\newblock In \emph{Proceedings of the 40th Annual Meeting of the Association for Computational Linguistics}, pages 311--318, Philadelphia, Pennsylvania, USA. Association for Computational Linguistics.

\bibitem[{Post(2018)}]{post-2018-call}
Matt Post. 2018.
\newblock \href {https://doi.org/10.18653/v1/W18-6319} {A call for clarity in reporting {BLEU} scores}.
\newblock In \emph{Proceedings of the Third Conference on Machine Translation: Research Papers}, pages 186--191, Brussels, Belgium. Association for Computational Linguistics.

\bibitem[{Smith(2007)}]{smith2007tesseractocr}
R.~Smith. 2007.
\newblock \href {https://doi.org/10.1109/ICDAR.2007.4376991} {An overview of the tesseract ocr engine}.
\newblock In \emph{Ninth International Conference on Document Analysis and Recognition (ICDAR 2007)}, volume~2, pages 629--633.

\bibitem[{Suzumura et~al.(2022)Suzumura, Sugiki, Takizawa, Imakura, Nakamura, Taura, Kudoh, Hanawa, Sekiya, Kobayashi, Kuga, Nakamura, Jiang, Kawase, Hanai, Miyazaki, Ishizaki, Shimotoku, Miyamoto, Aida, Takefusa, Kurimoto, Sasayama, Kitagawa, Fujiwara, Tanimura, Aoki, Endo, Ohshima, Fukazawa, Date, and Uchibayashi}]{9927975mdx}
Toyotaro Suzumura, Akiyoshi Sugiki, Hiroyuki Takizawa, Akira Imakura, Hiroshi Nakamura, Kenjiro Taura, Tomohiro Kudoh, Toshihiro Hanawa, Yuji Sekiya, Hiroki Kobayashi, Yohei Kuga, Ryo Nakamura, Renhe Jiang, Junya Kawase, Masatoshi Hanai, Hiroshi Miyazaki, Tsutomu Ishizaki, Daisuke Shimotoku, Daisuke Miyamoto, Kento Aida, Atsuko Takefusa, Takashi Kurimoto, Koji Sasayama, Naoya Kitagawa, Ikki Fujiwara, Yusuke Tanimura, Takayuki Aoki, Toshio Endo, Satoshi Ohshima, Keiichiro Fukazawa, Susumu Date, and Toshihiro Uchibayashi. 2022.
\newblock \href {https://doi.org/10.1109/DASC/PiCom/CBDCom/Cy55231.2022.9927975} {mdx: A cloud platform for supporting data science and cross-disciplinary research collaborations}.
\newblock In \emph{2022 IEEE Intl Conf on Dependable, Autonomic and Secure Computing, Intl Conf on Pervasive Intelligence and Computing, Intl Conf on Cloud and Big Data Computing, Intl Conf on Cyber Science and Technology Congress (DASC/PiCom/CBDCom/CyberSciTech)}, pages 1--7.

\bibitem[{Team et~al.(2025)Team, Kamath, Ferret, Pathak, Vieillard, Merhej, Perrin, Matejovicova, Ramé, Rivière, Rouillard, Mesnard, Cideron, bastien Grill, Ramos, Yvinec, Casbon, Pot, Penchev, Liu, Visin, Kenealy, Beyer, Zhai, Tsitsulin, Busa-Fekete, Feng, Sachdeva, Coleman, Gao, Mustafa, Barr, Parisotto, Tian, Eyal, Cherry, Peter, Sinopalnikov, Bhupatiraju, Agarwal, Kazemi, Malkin, Kumar, Vilar, Brusilovsky, Luo, Steiner, Friesen, Sharma, Sharma, Gilady, Goedeckemeyer, Saade, Feng, Kolesnikov, Bendebury, Abdagic, Vadi, György, Pinto, Das, Bapna, Miech, Yang, Paterson, Shenoy, Chakrabarti, Piot, Wu, Shahriari, Petrini, Chen, Lan, Choquette-Choo, Carey, Brick, Deutsch, Eisenbud, Cattle, Cheng, Paparas, Sreepathihalli, Reid, Tran, Zelle, Noland, Huizenga, Kharitonov, Liu, Amirkhanyan, Cameron, Hashemi, Klimczak-Plucińska, Singh, Mehta, Lehri, Hazimeh, Ballantyne, Szpektor, Nardini, Pouget-Abadie, Chan, Stanton, Wieting, Lai, Orbay, Fernandez, Newlan, yeong Ji, Singh, Black, Yu, Hui, Vodrahalli, Greff, Qiu,
  Valentine, Coelho, Ritter, Hoffman, Watson, Chaturvedi, Moynihan, Ma, Babar, Noy, Byrd, Roy, Momchev, Chauhan, Sachdeva, Bunyan, Botarda, Caron, Rubenstein, Culliton, Schmid, Sessa, Xu, Stanczyk, Tafti, Shivanna, Wu, Pan, Rokni, Willoughby, Vallu, Mullins, Jerome, Smoot, Girgin, Iqbal, Reddy, Sheth, Põder, Bhatnagar, Panyam, Eiger, Zhang, Liu, Yacovone, Liechty, Kalra, Evci, Misra, Roseberry, Feinberg, Kolesnikov, Han, Kwon, Chen, Chow, Zhu, Wei, Egyed, Cotruta, Giang, Kirk, Rao, Black, Babar, Lo, Moreira, Martins, Sanseviero, Gonzalez, Gleicher, Warkentin, Mirrokni, Senter, Collins, Barral, Ghahramani, Hadsell, Matias, Sculley, Petrov, Fiedel, Shazeer, Vinyals, Dean, Hassabis, Kavukcuoglu, Farabet, Buchatskaya, Alayrac, Anil, Dmitry, Lepikhin, Borgeaud, Bachem, Joulin, Andreev, Hardin, Dadashi, and Hussenot}]{gemmateam2025gemma3technicalreport}
Gemma Team, Aishwarya Kamath, Johan Ferret, Shreya Pathak, Nino Vieillard, Ramona Merhej, Sarah Perrin, Tatiana Matejovicova, Alexandre Ramé, Morgane Rivière, Louis Rouillard, Thomas Mesnard, Geoffrey Cideron, Jean bastien Grill, Sabela Ramos, Edouard Yvinec, Michelle Casbon, Etienne Pot, Ivo Penchev, Gaël Liu, Francesco Visin, Kathleen Kenealy, Lucas Beyer, Xiaohai Zhai, Anton Tsitsulin, Robert Busa-Fekete, Alex Feng, Noveen Sachdeva, Benjamin Coleman, Yi~Gao, Basil Mustafa, Iain Barr, Emilio Parisotto, David Tian, Matan Eyal, Colin Cherry, Jan-Thorsten Peter, Danila Sinopalnikov, Surya Bhupatiraju, Rishabh Agarwal, Mehran Kazemi, Dan Malkin, Ravin Kumar, David Vilar, Idan Brusilovsky, Jiaming Luo, Andreas Steiner, Abe Friesen, Abhanshu Sharma, Abheesht Sharma, Adi~Mayrav Gilady, Adrian Goedeckemeyer, Alaa Saade, Alex Feng, Alexander Kolesnikov, Alexei Bendebury, Alvin Abdagic, Amit Vadi, András György, André~Susano Pinto, Anil Das, Ankur Bapna, Antoine Miech, Antoine Yang, Antonia Paterson, Ashish
  Shenoy, Ayan Chakrabarti, Bilal Piot, Bo~Wu, Bobak Shahriari, Bryce Petrini, Charlie Chen, Charline~Le Lan, Christopher~A. Choquette-Choo, CJ~Carey, Cormac Brick, Daniel Deutsch, Danielle Eisenbud, Dee Cattle, Derek Cheng, Dimitris Paparas, Divyashree~Shivakumar Sreepathihalli, Doug Reid, Dustin Tran, Dustin Zelle, Eric Noland, Erwin Huizenga, Eugene Kharitonov, Frederick Liu, Gagik Amirkhanyan, Glenn Cameron, Hadi Hashemi, Hanna Klimczak-Plucińska, Harman Singh, Harsh Mehta, Harshal~Tushar Lehri, Hussein Hazimeh, Ian Ballantyne, Idan Szpektor, Ivan Nardini, Jean Pouget-Abadie, Jetha Chan, Joe Stanton, John Wieting, Jonathan Lai, Jordi Orbay, Joseph Fernandez, Josh Newlan, Ju~yeong Ji, Jyotinder Singh, Kat Black, Kathy Yu, Kevin Hui, Kiran Vodrahalli, Klaus Greff, Linhai Qiu, Marcella Valentine, Marina Coelho, Marvin Ritter, Matt Hoffman, Matthew Watson, Mayank Chaturvedi, Michael Moynihan, Min Ma, Nabila Babar, Natasha Noy, Nathan Byrd, Nick Roy, Nikola Momchev, Nilay Chauhan, Noveen Sachdeva, Oskar
  Bunyan, Pankil Botarda, Paul Caron, Paul~Kishan Rubenstein, Phil Culliton, Philipp Schmid, Pier~Giuseppe Sessa, Pingmei Xu, Piotr Stanczyk, Pouya Tafti, Rakesh Shivanna, Renjie Wu, Renke Pan, Reza Rokni, Rob Willoughby, Rohith Vallu, Ryan Mullins, Sammy Jerome, Sara Smoot, Sertan Girgin, Shariq Iqbal, Shashir Reddy, Shruti Sheth, Siim Põder, Sijal Bhatnagar, Sindhu~Raghuram Panyam, Sivan Eiger, Susan Zhang, Tianqi Liu, Trevor Yacovone, Tyler Liechty, Uday Kalra, Utku Evci, Vedant Misra, Vincent Roseberry, Vlad Feinberg, Vlad Kolesnikov, Woohyun Han, Woosuk Kwon, Xi~Chen, Yinlam Chow, Yuvein Zhu, Zichuan Wei, Zoltan Egyed, Victor Cotruta, Minh Giang, Phoebe Kirk, Anand Rao, Kat Black, Nabila Babar, Jessica Lo, Erica Moreira, Luiz~Gustavo Martins, Omar Sanseviero, Lucas Gonzalez, Zach Gleicher, Tris Warkentin, Vahab Mirrokni, Evan Senter, Eli Collins, Joelle Barral, Zoubin Ghahramani, Raia Hadsell, Yossi Matias, D.~Sculley, Slav Petrov, Noah Fiedel, Noam Shazeer, Oriol Vinyals, Jeff Dean, Demis Hassabis,
  Koray Kavukcuoglu, Clement Farabet, Elena Buchatskaya, Jean-Baptiste Alayrac, Rohan Anil, Dmitry, Lepikhin, Sebastian Borgeaud, Olivier Bachem, Armand Joulin, Alek Andreev, Cassidy Hardin, Robert Dadashi, and Léonard Hussenot. 2025.
\newblock \href {http://arxiv.org/abs/2503.19786} {Gemma 3 technical report}.

\bibitem[{Wei et~al.(2025)Wei, Sun, and Li}]{wei2025deepseekocrcontextsopticalcompression}
Haoran Wei, Yaofeng Sun, and Yukun Li. 2025.
\newblock \href {http://arxiv.org/abs/2510.18234} {Deepseek-ocr: Contexts optical compression}.

\bibitem[{Xu et~al.(2021)Xu, Xu, Lv, Cui, Wei, Wang, Lu, Florencio, Zhang, Che, Zhang, and Zhou}]{xu-etal-2021-layoutlmv2}
Yang Xu, Yiheng Xu, Tengchao Lv, Lei Cui, Furu Wei, Guoxin Wang, Yijuan Lu, Dinei Florencio, Cha Zhang, Wanxiang Che, Min Zhang, and Lidong Zhou. 2021.
\newblock \href {https://doi.org/10.18653/v1/2021.acl-long.201} {{L}ayout{LM}v2: Multi-modal pre-training for visually-rich document understanding}.
\newblock In \emph{Proceedings of the 59th Annual Meeting of the Association for Computational Linguistics and the 11th International Joint Conference on Natural Language Processing (Volume 1: Long Papers)}, pages 2579--2591, Online. Association for Computational Linguistics.

\bibitem[{Xu et~al.(2020)Xu, Li, Cui, Huang, Wei, and Zhou}]{10.1145/3394486.3403172}
Yiheng Xu, Minghao Li, Lei Cui, Shaohan Huang, Furu Wei, and Ming Zhou. 2020.
\newblock \href {https://doi.org/10.1145/3394486.3403172} {Layoutlm: Pre-training of text and layout for document image understanding}.
\newblock In \emph{Proceedings of the 26th ACM SIGKDD International Conference on Knowledge Discovery \& Data Mining}, KDD '20, page 1192–1200, New York, NY, USA. Association for Computing Machinery.

\bibitem[{Ye et~al.(2023{\natexlab{a}})Ye, Hu, Xu, Ye, Yan, Dan, Zhao, Xu, Li, Tian, Qi, Zhang, and Huang}]{ye2023mplugdocowlmodularizedmultimodallarge}
Jiabo Ye, Anwen Hu, Haiyang Xu, Qinghao Ye, Ming Yan, Yuhao Dan, Chenlin Zhao, Guohai Xu, Chenliang Li, Junfeng Tian, Qian Qi, Ji~Zhang, and Fei Huang. 2023{\natexlab{a}}.
\newblock \href {http://arxiv.org/abs/2307.02499} {mplug-docowl: Modularized multimodal large language model for document understanding}.

\bibitem[{Ye et~al.(2023{\natexlab{b}})Ye, Hu, Xu, Ye, Yan, Xu, Li, Tian, Qian, Zhang, Jin, He, Lin, and Huang}]{ye-etal-2023-ureader}
Jiabo Ye, Anwen Hu, Haiyang Xu, Qinghao Ye, Ming Yan, Guohai Xu, Chenliang Li, Junfeng Tian, Qi~Qian, Ji~Zhang, Qin Jin, Liang He, Xin Lin, and Fei Huang. 2023{\natexlab{b}}.
\newblock \href {https://doi.org/10.18653/v1/2023.findings-emnlp.187} {{UR}eader: Universal {OCR}-free visually-situated language understanding with multimodal large language model}.
\newblock In \emph{Findings of the Association for Computational Linguistics: EMNLP 2023}, pages 2841--2858, Singapore. Association for Computational Linguistics.

\bibitem[{Zhu et~al.(2025)Zhu, Wang, Chen, Liu, Ye, Gu, Tian, Duan, Su, Shao, Gao, Cui, Wang, Cao, Liu, Wei, Zhang, Wang, Xu, Li, Wang, Deng, Li, He, Jiang, Luo, Wang, He, Shi, Zhang, Shao, He, Xiong, Qu, Sun, Jiao, Lv, Wu, Zhang, Deng, Ge, Chen, Wang, Dou, Lu, Zhu, Lu, Lin, Qiao, Dai, and Wang}]{zhu2025internvl3exploringadvancedtraining}
Jinguo Zhu, Weiyun Wang, Zhe Chen, Zhaoyang Liu, Shenglong Ye, Lixin Gu, Hao Tian, Yuchen Duan, Weijie Su, Jie Shao, Zhangwei Gao, Erfei Cui, Xuehui Wang, Yue Cao, Yangzhou Liu, Xingguang Wei, Hongjie Zhang, Haomin Wang, Weiye Xu, Hao Li, Jiahao Wang, Nianchen Deng, Songze Li, Yinan He, Tan Jiang, Jiapeng Luo, Yi~Wang, Conghui He, Botian Shi, Xingcheng Zhang, Wenqi Shao, Junjun He, Yingtong Xiong, Wenwen Qu, Peng Sun, Penglong Jiao, Han Lv, Lijun Wu, Kaipeng Zhang, Huipeng Deng, Jiaye Ge, Kai Chen, Limin Wang, Min Dou, Lewei Lu, Xizhou Zhu, Tong Lu, Dahua Lin, Yu~Qiao, Jifeng Dai, and Wenhai Wang. 2025.
\newblock \href {http://arxiv.org/abs/2504.10479} {Internvl3: Exploring advanced training and test-time recipes for open-source multimodal models}.

\end{thebibliography}

\section{Language Resource References}
\label{lr:ref}
\bibliographystylelanguageresource{lrec2026-natbib}
\bibliographylanguageresource{languageresource}

\section{Appendix}

\subsection{Evaluation of OCR-Specialized MLLMs}

We evaluated two recent OCR-specialized MLLMs, DeepSeek-OCR~\cite{wei2025deepseekocrcontextsopticalcompression} and PaddleOCR-VL~\cite{cui2025paddleocrvlboostingmultilingualdocument}, on our proposed dataset.
For DeepSeek-OCR, we employed the Gundam (Dynamic Resolution) mode with the prompt ``\verb|<image>\nFree OCR. |''.
The max new tokens parameter was set to 1024 for JSSODa and 3072 for VJRODa.
For PaddleOCR-VL, we adopted a pipeline in which document layout analysis was first performed and the resulting segments were fed to the MLLM; the final answer was obtained by concatenating the text according to the predicted reading order.
No max new tokens constraint was specified for PaddleOCR-VL.
Note that these settings, including the prompt and max new tokens configurations, differ from those used for other models.

Table~\ref{table:eval_jssoda_ocr_mllm} presents the results on the JSSODa test set.
DeepSeek-OCR exhibited strong text recognition performance for horizontally written text but performed poorly on vertically written text.
In contrast, PaddleOCR-VL showed much less degradation on vertically written text than DeepSeek-OCR.
Table~\ref{table:eval_rw_ocr_mllm} reports the results on the VJRODa dataset.
DeepSeek-OCR obtained low scores, indicating weak performance on vertically written Japanese text.
Across all models, including those listed in Table~\ref{table:eval_rw}, PaddleOCR-VL achieved the highest scores.

\begin{table*}[t]
\centering
\footnotesize
\begin{tabular}{lcccccccc}
\toprule
\textbf{Columns} & \multicolumn{2}{c}{1} & \multicolumn{2}{c}{2} & \multicolumn{2}{c}{3} & \multicolumn{2}{c}{4} \\
\cmidrule(lr){1-1} \cmidrule(lr){2-3} \cmidrule(lr){4-5} \cmidrule(lr){6-7} \cmidrule(lr){8-9}
\textbf{Models} & CER($\downarrow$) & BLEU($\uparrow$) & CER($\downarrow$) & BLEU($\uparrow$) & CER($\downarrow$) & BLEU($\uparrow$) & CER($\downarrow$) & BLEU($\uparrow$) \\
\midrule
\midrule
\multicolumn{9}{c}{\textbf{horizontal Writing}} \\
\midrule
\multicolumn{9}{c}{Raw Output} \\
\midrule
DeepSeek-OCR  & 0.190 & 99.6 & 6.09 & 99.3 & 3.08 & 99.0 & 3.10 & 96.7 \\
PaddleOCR-VL  & 0.496 & 99.1 & 3.58 & 96.7 & 12.6 & 93.7 & 18.5 & 90.4 \\
\midrule
\multicolumn{9}{c}{Remove Repetition} \\
\midrule
DeepSeek-OCR  & 0.190 & 99.6 & 6.09 & 99.3 & 3.08 & 99.0 & 3.10 & 96.7 \\
PaddleOCR-VL  & 0.496 & 99.1 & 3.58 & 96.7 & 12.6 & 93.7 & 18.5 & 90.4 \\
\midrule
\midrule
\multicolumn{9}{c}{\textbf{Vertical Writing}} \\
\midrule
\multicolumn{9}{c}{Raw Output} \\
\midrule
DeepSeek-OCR  & 108 & 35.7 & 160 & 11.7 & 153 & 12.4 & 130 & 9.80 \\
PaddleOCR-VL  & 27.3 & 97.2 & 11.0 & 90.4 & 7.75 & 93.0 & 6.67 & 94.1 \\
\midrule
\multicolumn{9}{c}{Remove Repetition} \\
\midrule
DeepSeek-OCR  & 108 & 35.8 & 158 & 11.8 & 152 & 12.6 & 129 & 9.87 \\
PaddleOCR-VL  & 27.2 & 97.2 & 3.75 & 96.4 & 4.61 & 96.1 & 4.00 & 96.5 \\
\bottomrule
\end{tabular}
\caption{The result on JSSODa test set}
\label{table:eval_jssoda_ocr_mllm}
\end{table*}

\begin{table*}[t]
\centering
\footnotesize
\begin{tabular}{lcccc}
\toprule
 & \multicolumn{2}{c}{Raw Output} & \multicolumn{2}{c}{Remove Repetition} \\
 \cmidrule(lr){2-3} \cmidrule(lr){4-5}
\textbf{Models} & CER($\downarrow$) & BLEU($\uparrow$) & CER($\downarrow$) & BLEU($\uparrow$) \\
\midrule
DeepSeek-OCR             & 182  & 10.8 & 181  & 10.9 \\ 
PaddleOCR-VL             & 20.1 & 91.4 & 19.1 & 91.3  \\ 
\bottomrule
\end{tabular}
\caption{The result on VJRODa}
\label{table:eval_rw_ocr_mllm}
\end{table*}

\subsection{Prompts}

Table~\ref{table:gen_sentences_prompt} shows the prompt used to instruct the LLM to generate Japanese sentences from Japanese nouns during the construction of the JSSODa dataset.
Table~\ref{table:inference_prompt} shows the user prompt used for fine-tuning and evaluation.

\begin{table*}[t]
\centering
\footnotesize
\begin{tabularx}{\linewidth}{@{}X@{}}
\toprule
\begin{minipage}{\linewidth}
以下の単語について、その単語をテーマにした日本語の文章を出力してください。文章は500文字以上にしてください。与えられた単語が必ずしも出力に含まれている必要はありません。文章の一部に英単語や数字が含まれていてもよいです。文章の文体はどのようなものでもよく、教科書風の文章、ニュース記事、小説、エッセイ、プレスリリース、官公庁の文章、SNSなど、日本語として破綻していなければ何でもよいです。出力は文章のみとし、余計なものは出力しないでください。
\\ \\
単語: \{word\}
\\ \\
\textcolor{red}{(\textit{For the following word, output a Japanese text themed around that word. Make the text at least 500 characters long. The given word does not necessarily need to be included in the output. It is acceptable for parts of the text to contain English words or numbers. Any writing style is fine—textbook-like writing, a news article, a novel, an essay, a press release, an official government-style document, social media, etc.—as long as it is coherent in Japanese. Output text only, and do not include anything unnecessary.
\\ \\
Word: \{word\}})}
\end{minipage} \\
\bottomrule
\end{tabularx}
\caption{Prompt for generating Japanese sentences from Japanese nouns}
\label{table:gen_sentences_prompt}
\end{table*}

\begin{table*}[t]
\centering
\footnotesize
\begin{tabularx}{\linewidth}{@{}X@{}}
\toprule
\begin{minipage}{\linewidth}
この画像内のテキストを日本語の読み順に従って全て出力してください。出力は画像内のテキストのみとしてください。
\\ \\
\textcolor{red}{(\textit{Please output all the text in this image following the Japanese reading order. The output should contain only the text in the image.})}
\end{minipage} \\
\bottomrule
\end{tabularx}
\caption{Prompt for fine-tuining and evaluation}
\label{table:inference_prompt}
\end{table*}

\subsection{Training Budget}
When trained on our JSSODa train set, Qwen2.5-VL-7B-Instruct requires 2.5 hours on 1 node with 8 × NVIDIA A100 (40 GB) GPUs; InternVL3-8B-hf requires 2 hours on 2 nodes, each with 8 × NVIDIA A100 (40 GB) GPUs; and Gemma 3 12B IT requires 1 hour on 2 nodes, each with 8 × NVIDIA A100 (40 GB) GPUs.

\subsection{Regular Expression for Removing Repetition}
The regular expressions used to remove repeated segments from LLM-generated text are presented in Table~\ref{table:regular_expression}.

\begin{table*}[t]
\centering
\footnotesize
\begin{tabularx}{\linewidth}{@{}X@{}}
\toprule
\begin{minipage}{\linewidth}
\verb|repetition_pattern = r'((\S.*?)\2{9,})(?!.*(\S.*?)\3{9,})'|

\verb|pred_text = re.sub(repetition_pattern, r'\2', pred_text, flags=re.DOTALL)|
\end{minipage} \\
\bottomrule
\end{tabularx}
\caption{Regular expression for removing repetition}
\label{table:regular_expression}
\end{table*}

\subsection{API Versions}
The API versions used were \texttt{gpt-4.1-2025-04-14} for GPT-4.1 and \texttt{gpt-5-2025-08-07} for GPT-5.

\subsection{Additional Examples of Images from Our Datasets}
\begin{figure*}[t]
  \begin{minipage}[b]{\columnwidth}
    \centering
    \includegraphics[keepaspectratio, width=\columnwidth]{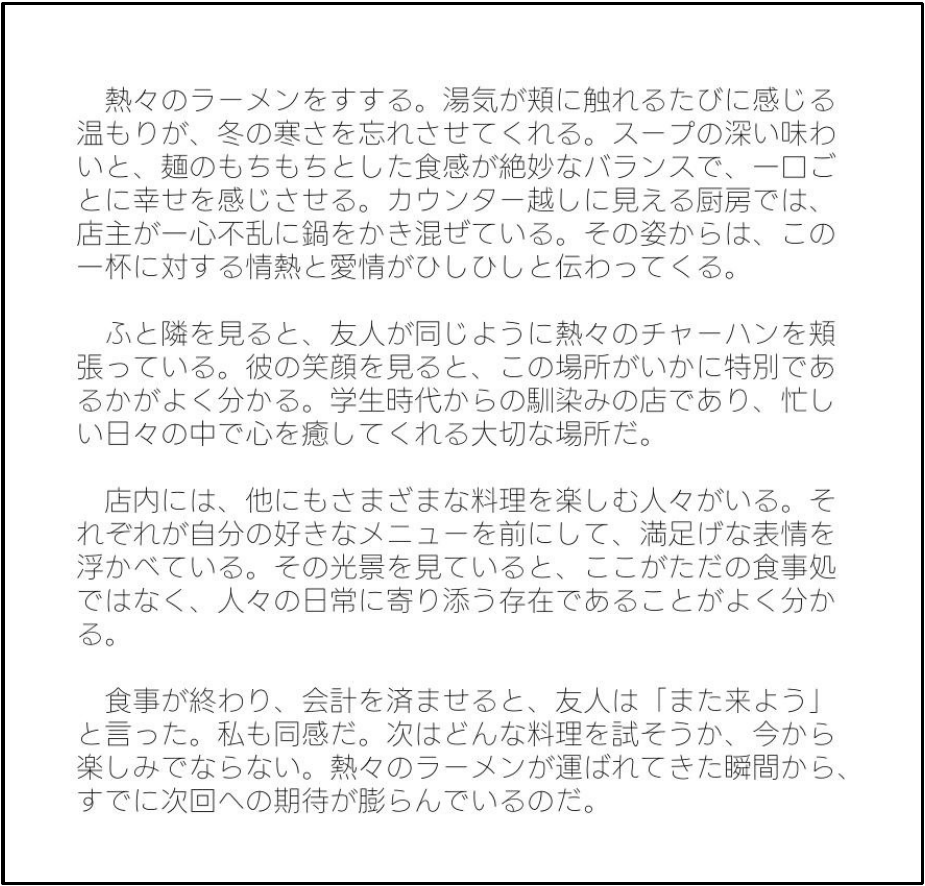}
    \subcaption{horizontal, 1-column}
  \end{minipage}
  \begin{minipage}[b]{\columnwidth}
    \centering
    \includegraphics[keepaspectratio, width=\columnwidth]{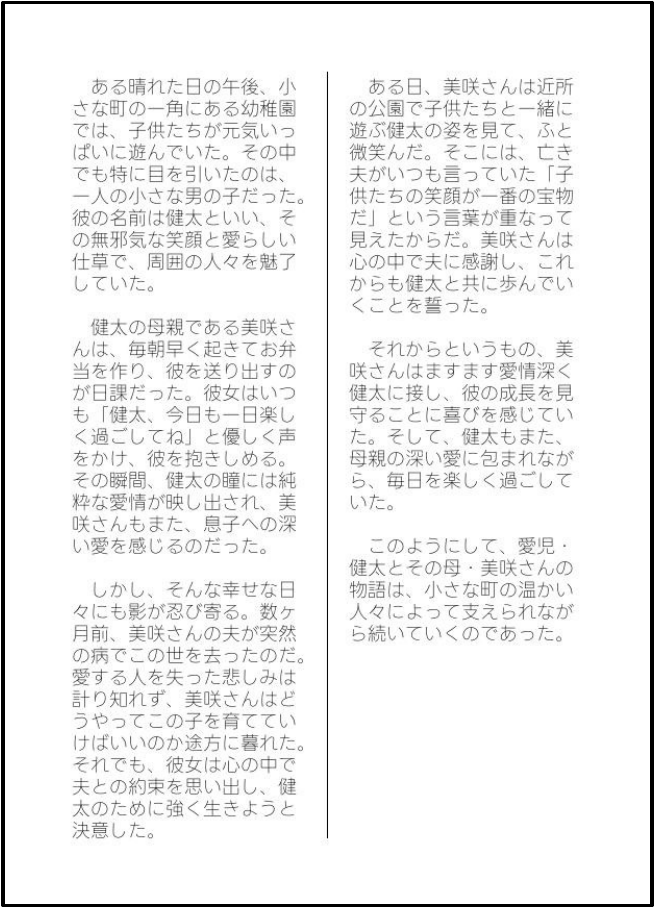}
    \subcaption{horizontal, 2-columns}
  \end{minipage} \\
  \begin{minipage}[b]{\columnwidth}
    \centering
    \includegraphics[keepaspectratio, width=\columnwidth]{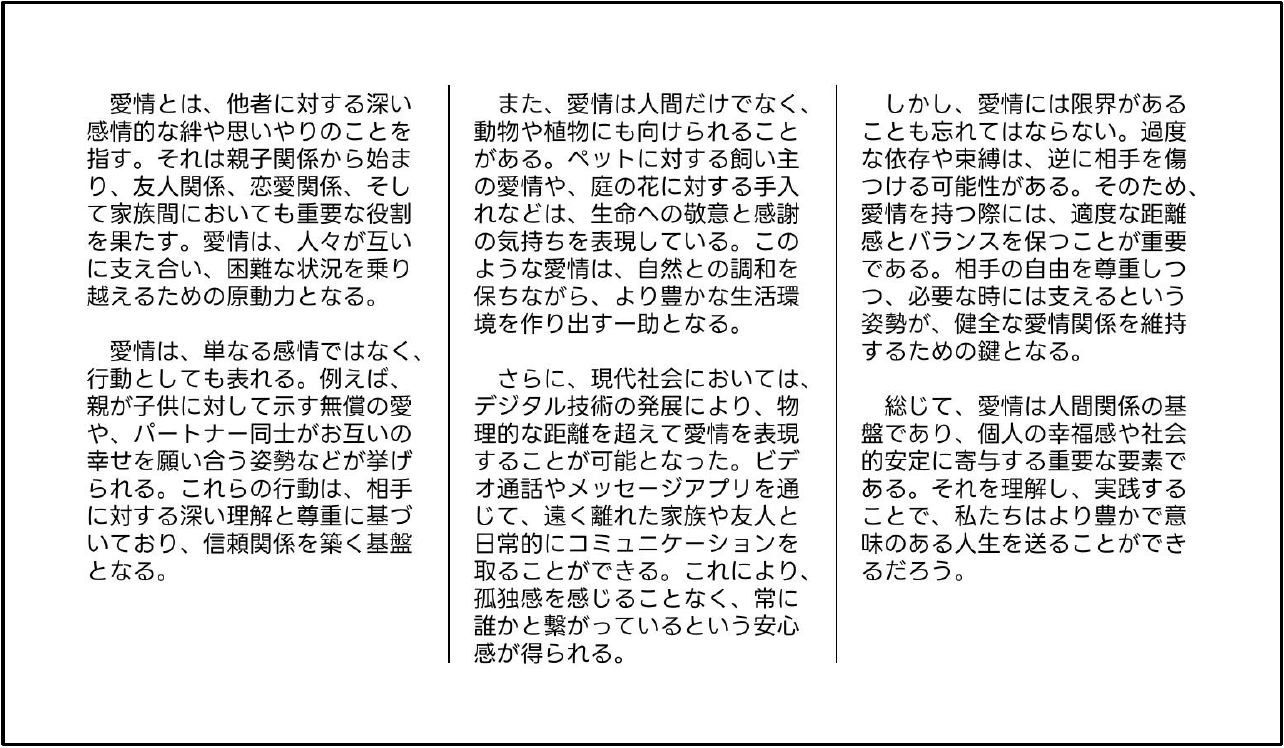}
    \subcaption{horizontal, 3-columns}
  \end{minipage} 
  \begin{minipage}[b]{\columnwidth}
    \centering
    \includegraphics[keepaspectratio, width=\columnwidth]{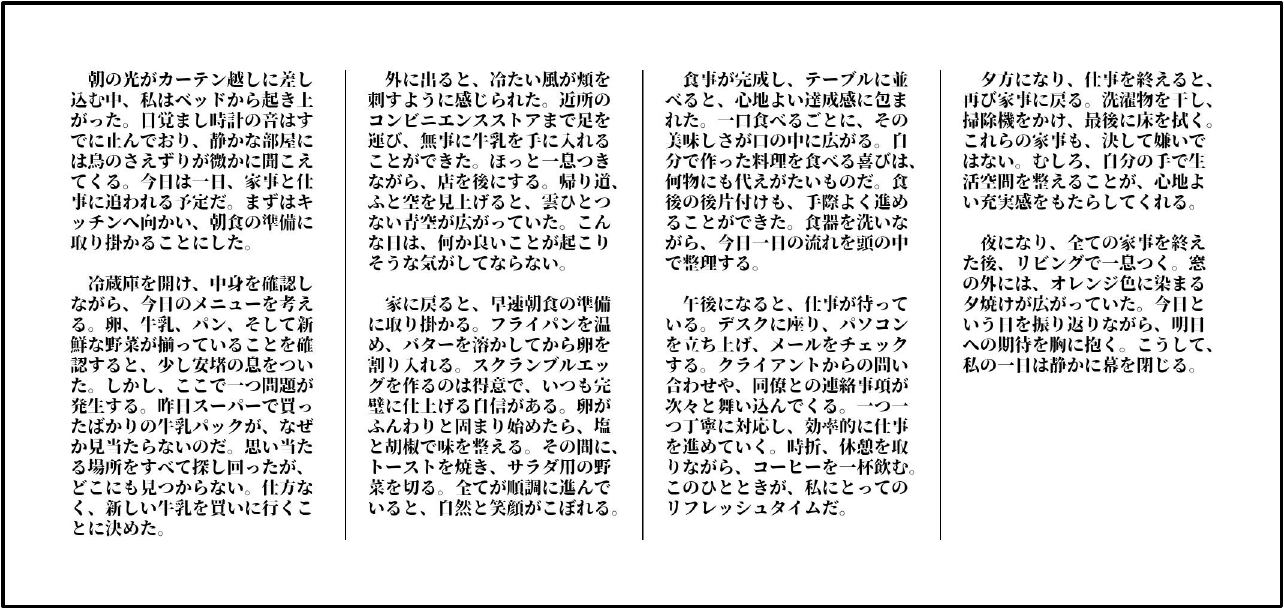}
    \subcaption{horizontal, 4-columns}
  \end{minipage} \\
  \caption{Additional example images from JSSODa (horizontal writing)}
  \label{fig:appendix_jssoda_h}
\end{figure*}

\begin{figure*}[t]
  \begin{minipage}[b]{\columnwidth}
    \centering
    \includegraphics[keepaspectratio, width=0.66\columnwidth]{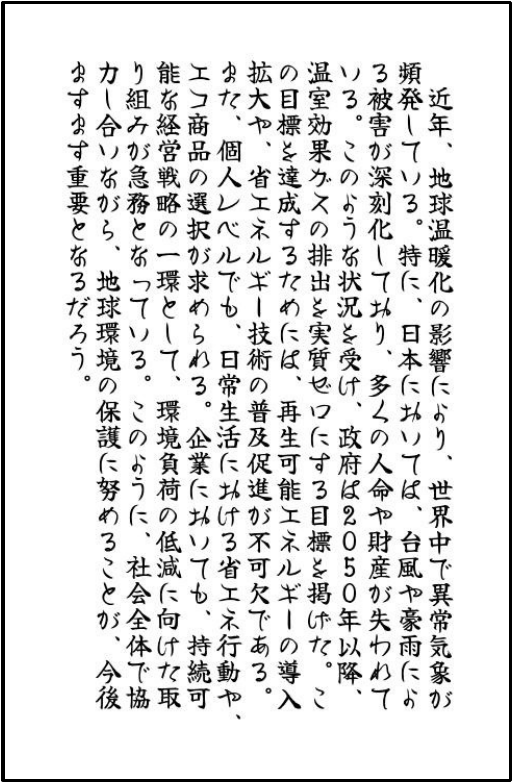}
    \subcaption{vertical, 1-column}
  \end{minipage}
  \begin{minipage}[b]{\columnwidth}
    \centering
    \includegraphics[keepaspectratio, width=\columnwidth]{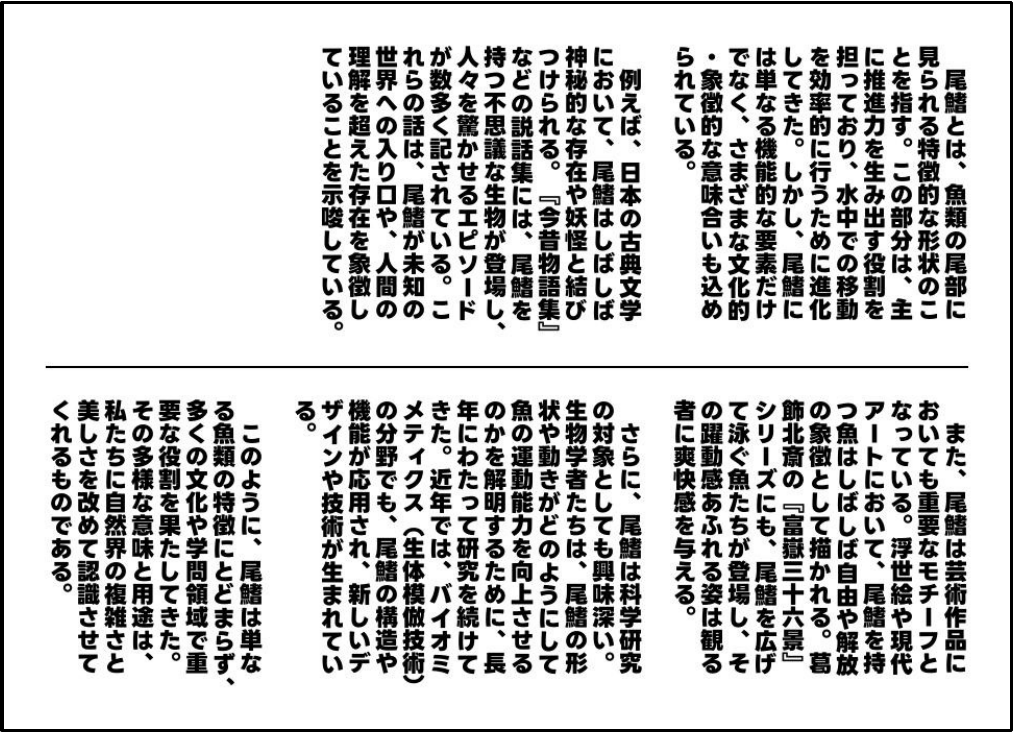}
    \subcaption{vertical, 2-columns}
  \end{minipage} \\
  \begin{minipage}[b]{\columnwidth}
    \centering
    \includegraphics[keepaspectratio, width=\columnwidth]{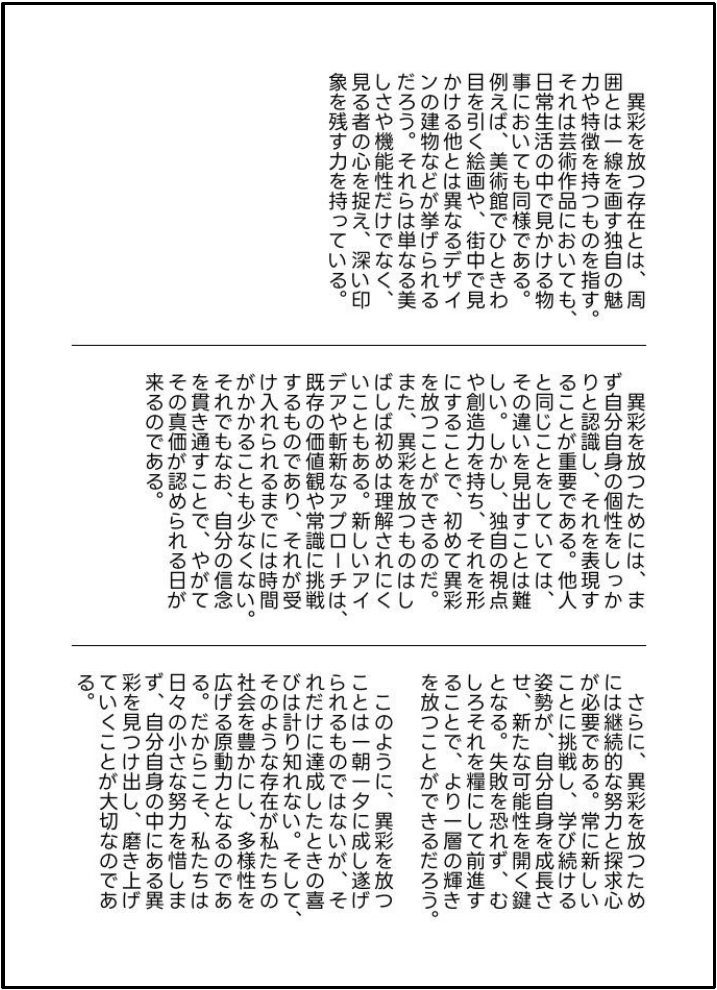}
    \subcaption{vertical, 3-columns}
  \end{minipage}
  \begin{minipage}[b]{\columnwidth}
    \centering
    \includegraphics[keepaspectratio, width=0.66\columnwidth]{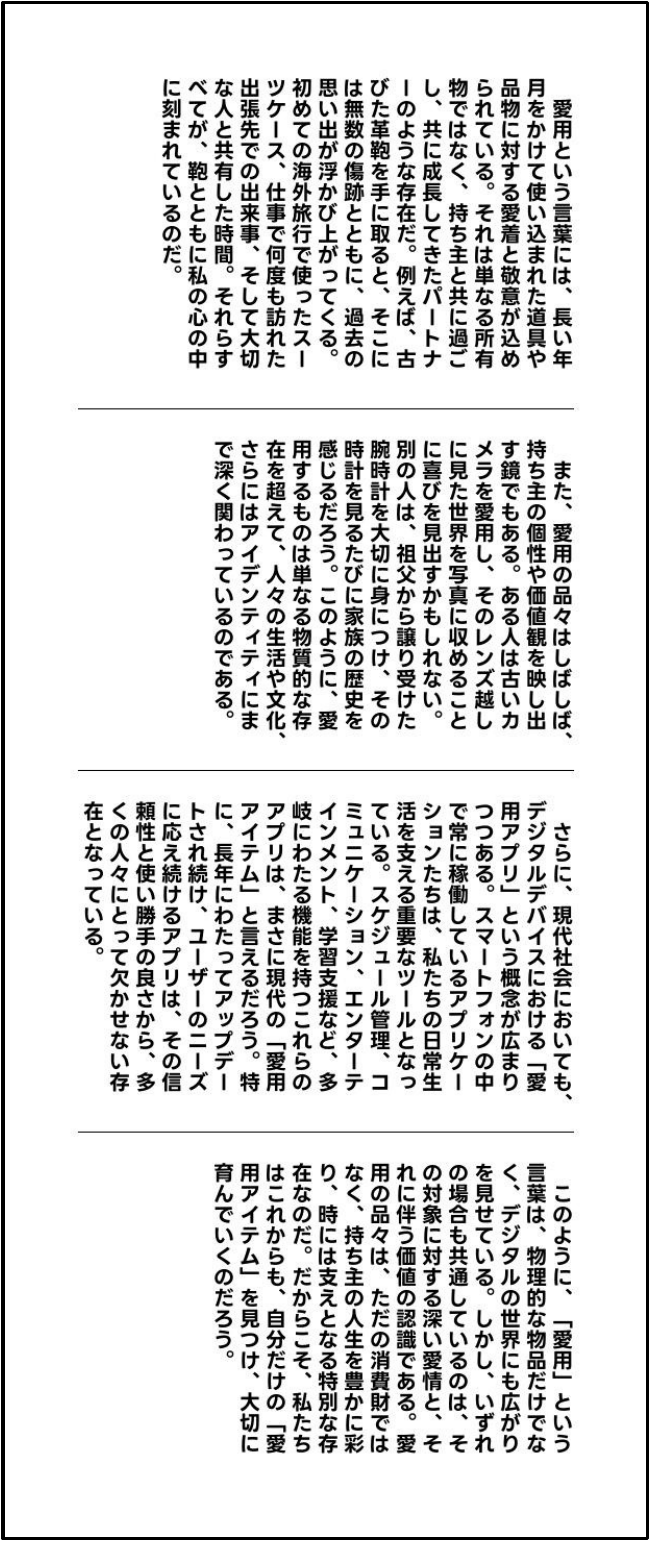}
    \subcaption{vertical, 4-columns}
  \end{minipage} \\
  \caption{Additional example images from JSSODa (vertical writing)}
  \label{fig:appendix_jssoda_v}
\end{figure*}

\begin{figure*}[t]
  \begin{minipage}[b]{\columnwidth}
    \centering
    \includegraphics[keepaspectratio, width=\columnwidth]{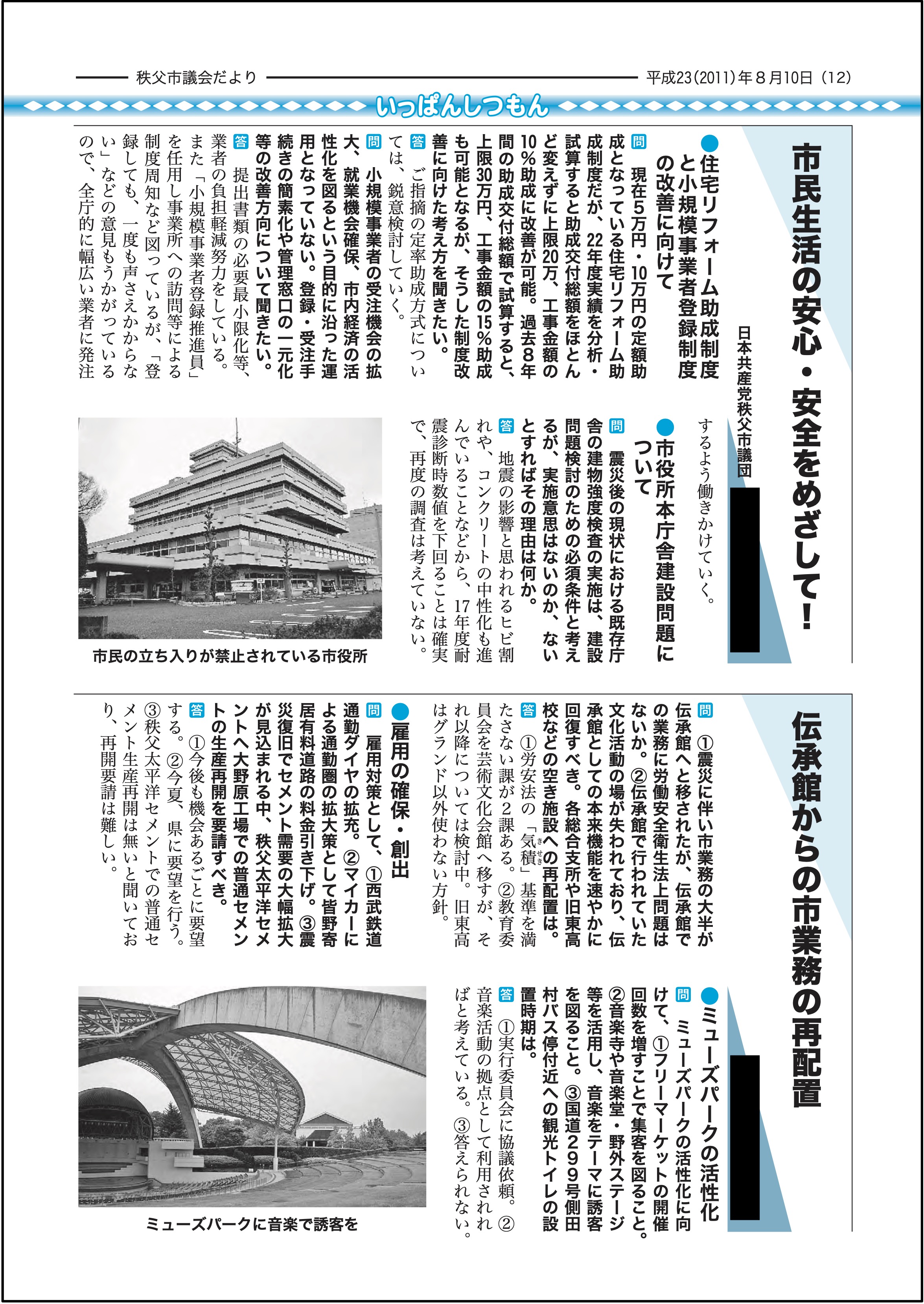}
  \end{minipage}
  \begin{minipage}[b]{\columnwidth}
    \centering
    \includegraphics[keepaspectratio, width=\columnwidth]{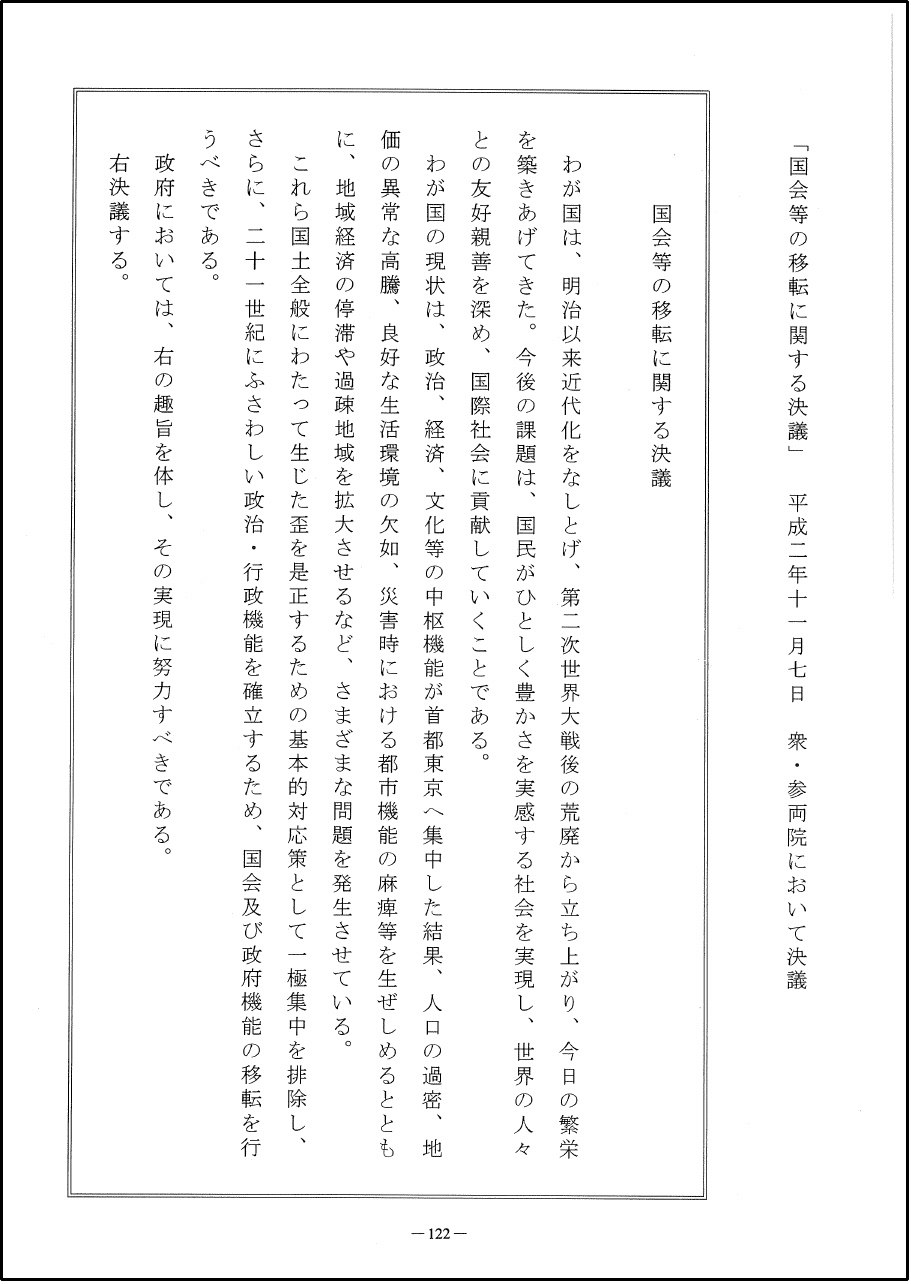}
  \end{minipage} \\
  \begin{minipage}[b]{\linewidth}
    \centering
    \includegraphics[keepaspectratio, width=\linewidth]{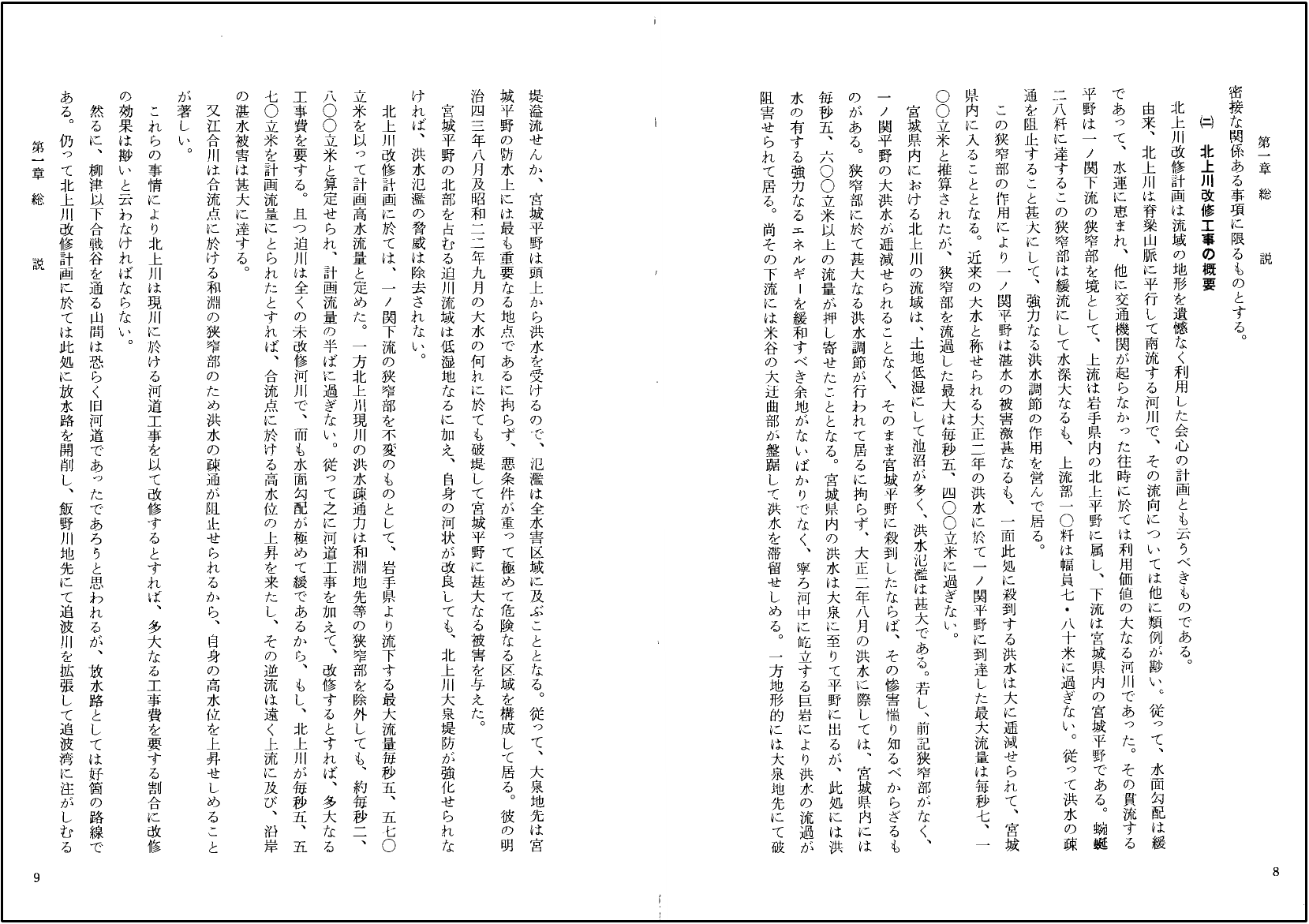}
  \end{minipage}
  \caption{
    Additional example images from VJRODa
    (top left: \protect\url{https://www.city.chichibu.lg.jp/secure/5707/\%E5\%B8\%82\%E8\%AD\%B0\%E4\%BC\%9A\%E3\%81\%A0\%E3\%82\%88\%E3\%82\%8A25\%E5\%8F\%B712-15.pdf}, page 1,
    top right: \protect\url{https://www.mlit.go.jp/kokudokeikaku/iten/information/council/pdf/toushin_sankou.pdf}, page 120,
    bottom: \protect\url{https://www.thr.mlit.go.jp/Bumon/J73101/homepage/iport/kitakami/asobu\_manabu/kitakamigawa/image/kitakami\_3.pdf}, page 14,
    Personal information has been masked.)
    }
  \label{fig:appendix_vjroda}
\end{figure*}

We show additional image examples from our JSSODa dataset in Figures~\ref{fig:appendix_jssoda_h} and~\ref{fig:appendix_jssoda_v}, and from our VJRODa dataset in Figure~\ref{fig:appendix_vjroda}.

\end{document}